\documentclass[lettersize,journal]{IEEEtran}
\usepackage{float}

\usepackage{cite}
\usepackage[numbers,compress]{natbib}
\usepackage{graphicx}
\usepackage{booktabs}
\usepackage{amssymb}
\usepackage[table]{xcolor}
\usepackage{tikz}
\usetikzlibrary{positioning, arrows.meta, calc}
\usepackage{svg}

\ifCLASSINFOpdf
\else
\fi

\usepackage{amsmath}
\usepackage{graphicx}
\usepackage{algorithmic}
\usepackage{array}
\usepackage{fixltx2e}
\usepackage{stfloats}
\usepackage{url}
\usepackage{subcaption}

\begin{document}

\title{BuildMamba: A Visual State-Space Based Model for Multi-Task Building Segmentation
and Height Estimation from Satellite Images}

\author{Sinan~U.~Ulu, A.~Enes~Doruk, I.~Can~Yagmur, Bahadir~K.~Gunturk, \\ Oguz~Hanoglu~and~Hasan~F.~Ates,~\IEEEmembership{Senior Member, IEEE}
\thanks{Sinan U. Ulu, A. Enes Doruk  and  Hasan F. Ates are  with the Graduate School of Engineering, Ozyegin university, 34794 Istanbul, Türkiye. (e-mail: hasan.ates@ozyegin.edu.tr).} a space
\thanks{I. Can Yagmur is with the Department of Computer Science, University of Rochester, Rochester, NY 14627 USA.}
\thanks{Bahadir K. Gunturk is with the School of Engineering and Natural Sciences, Istanbul Medipol University, 34810 Istanbul, Türkiye.}
\thanks{Oguz Hanoglu is with Huawei Turkiye R\&D Center, 34768 Istanbul, Türkiye.} 
}

\markboth{Journal of \LaTeX\ Class Files,~Vol.~?, No.~?, March~2026}%
{Shell \MakeLowercase{\textit{et al.}}: Bare Demo of IEEEtran.cls for IEEE Journals}

\maketitle

\begin{abstract}
Accurate building segmentation and height estimation from single-view RGB satellite imagery are fundamental for urban analytics, yet remain ill-posed due to structural variability and the high computational cost of global context modeling. While current approaches typically adapt monocular depth architectures, they often suffer from boundary bleeding and systematic underestimation of high-rise structures. To address these limitations, we propose BuildMamba, a unified multi-task framework designed to exploit the linear-time global modeling of visual state-space models. Motivated by the need for stronger structural coupling and computational efficiency, we introduce three modules: a Mamba Attention Module for dynamic spatial recalibration, a Spatial-Aware Mamba-FPN for multi-scale feature aggregation via gated state-space scans, and a Mask-Aware Height Refinement module using semantic priors to suppress height artifacts. Extensive experiments demonstrate that BuildMamba establishes a new performance upper bound across three benchmarks. Specifically, it achieves an IoU of 0.93 and RMSE of 1.77~m on DFC23 benchmark, surpassing state-of-the-art by 0.82~m in height estimation. Simulation results confirm the model's superior robustness and scalability for large-scale 3D urban reconstruction.
\end{abstract}


\IEEEpeerreviewmaketitle

\section{Introduction}

Computer vision has advanced rapidly, enabling large-scale analysis of satellite imagery and raising expectations for urban analytics. In this context, building footprints and heights constitute fundamental variables for numerous downstream applications. They support urban planning and land-use analysis by characterizing building density and spatial organization \cite{li2016urban}, infrastructure monitoring and large-scale assessment of the built environment using building inventories and 3D city models \cite{ijgi4042842,sirko2021openbuildings}, and risk and vulnerability assessment through the quantification of hazard exposure based on building geometry \cite{schroeter2018flood}. Building height and footprint information are also essential for shadow modeling and urban climate analysis \cite{urbanmorph2022}, communication network design and line-of-sight modeling \cite{alhourani2014uav}, and high-resolution population and asset mapping at regional and continental scales \cite{oostwegel2025openbuildingmap,eubucco2023}.

\begin{figure}[ht]
  \centering
  \includegraphics[width=\linewidth]{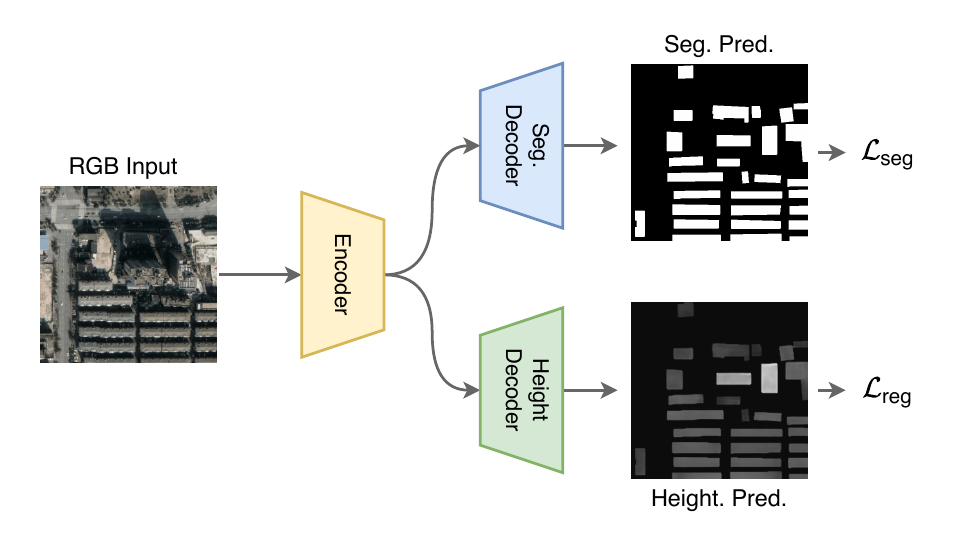}
   \caption{Overview of the proposed BuildMamba architecture for joint building segmentation and height estimation.}
  \label{fig:pipeline}
\end{figure}

Satellite-based remote sensing is uniquely suited to provide such information at regional and global scales, but it operates under a fundamentally different imaging geometry than ground-level vision. From a near top-down viewpoint, buildings appear primarily as two-dimensional radiometric patterns rather than explicit three-dimensional structures. This requires structural attributes such as height and precise footprint boundaries to be inferred indirectly from appearance cues, including texture, tone, and spatial context. As a result, joint building segmentation and height estimation from single-view RGB satellite imagery is inherently ill-posed and highly sensitive to illumination changes, shadows, atmospheric effects, roof materials, and heterogeneous urban layouts, leading to ambiguity and instability widely observed in monocular height estimation studies \cite{IM2HEIGHT, IM2ELEVATION, BuildFormer}.

Building height estimation from monocular satellite imagery is commonly addressed by adapting monocular depth estimation architectures, including encoder–decoder CNNs \cite{IM2HEIGHT, IM2ELEVATION}, ordinal-regression models \cite{dorn}, and building-specific transformer-based predictors \cite{BuildFormer}. Recent refinements incorporate adaptive binning strategies \cite{Adabins, binsformer} and hybrid classification-regression formulations \cite{chen2023htc} to mitigate ambiguities in monocular height retrieval and better handle long-tailed height distributions. Despite these improvements, most RGB-only approaches remain appearance-driven and rely on implicit correlations between image texture and height, with limited structural understanding of building shapes. Consequently, height prediction becomes highly sensitive to homogeneous rooftop textures, shadows, and radiometric distortions, frequently producing boundary artifacts, systematic underestimation of tall buildings, and limited generalization across cities and sensors. These issues are further exacerbated by noise in DSM (Digital Surface Model) or LiDAR-derived supervision caused by resolution mismatch, temporal misalignment, and registration errors.

Some works attempt to mitigate these limitations by incorporating additional modalities such as SAR \cite{mftsc,tridentmtl,sardatafusion} or auxiliary supervision such as land-use masks \cite{lumnet}. However, such information is frequently unavailable at inference time, restricting practical deployment in large-scale RGB-only settings. Moreover, in many joint segmentation–height estimation frameworks \cite{popnet, dsmnet}, height estimation is treated as a secondary output rather than a structurally constrained task, leaving the dependency between building footprints and heights under-exploited. These observations suggest that robust monocular height estimation requires both strong global contextual reasoning and precise local spatial modeling, together with explicit structural coupling between segmentation and height prediction.

Motivated by these limitations, recent remote sensing frameworks increasingly adopt dual-path encoders that separate global contextual modeling from local spatial feature extraction \cite{ST-Unet, FarSeg++, PointFlow}. Such architectures aim to reconcile the need for fine-grained spatial detail with long-range contextual reasoning, benefiting both building footprint extraction \cite{uanet} and joint segmentation–height estimation tasks \cite{popnet, light}. Global pathways often rely on transformer-based encoders to model scene-level context, while local pathways employ convolutional layers to preserve spatial locality and edge structure. 

More recently, state-space-driven visual models such as VMamba \cite{vmamba} have emerged as efficient alternatives to attention-based transformers \cite{vit} and hierarchical variants such as SwinTransformer \cite{swin}. By enabling long-range contextual modeling with linear computational complexity while preserving spatial locality through structured recurrence, VMamba offers a favorable balance between global reasoning and efficiency for high-resolution remote sensing imagery.

In parallel, multi-task and multi-modal formulations jointly learn building segmentation and height estimation using complementary data sources beyond RGB. DSMNet \cite{dsmnet} employs a two-stage design for joint prediction and refinement, while MFTSC \cite{mftsc} integrates transformer-based encoding with local feature extraction and pyramid pooling to impose semantic constraints on height regression. Joint contrastive learning approaches \cite{jointcontrastive} further promote feature consistency between segmentation and height prediction. Despite these advances, existing methods often rely on auxiliary modalities, rigid fusion strategies, or benchmark-specific assumptions, and remain sensitive to domain shifts caused by heterogeneous sensors and geographic variability. Moreover, height prediction is frequently weakly coupled to explicit building footprints and boundaries, limiting geometric consistency and robustness in practical RGB-only settings.

As a result, several core challenges remain unresolved in monocular satellite-based building height estimation. Existing methods struggle with tall-building prediction due to severe data imbalance and the scarcity of high-rise examples, leading to systematic underestimation. Boundary ambiguity, shadows, and structural variability introduce noise around complex roof geometries, while domain shifts across cities and sensors degrade generalization. Furthermore, reliance on auxiliary modalities limits scalability in practical RGB-only deployments. Addressing these limitations requires architectures that jointly model long-range context and fine-grained spatial detail, while explicitly coupling segmentation and height estimation.

To address these limitations, we propose BuildMamba, a unified multi-task framework for joint building segmentation and height estimation from single RGB satellite images. As summarized in Figure~\ref{fig:pipeline}, BuildMamba employs a global--local encoder built upon a visual state-space backbone to capture long-range contextual dependencies while preserving fine-grained spatial detail, followed by task-specific decoders for segmentation and height prediction. By explicitly enforcing structural coupling between building footprints and height regression, the framework enables more stable predictions, sharper boundaries, and improved robustness in RGB-only settings.

This article is a significantly extended version of our previous work \cite{buildmamba_conference}. While the preliminary framework established the feasibility of using a Visual State-Space backbone for joint segmentation and height estimation, the present work substantially strengthens architectural design and structural coupling. First, we propose a novel Spatial-Aware Mamba-FPN (Feature Pyramid Network) that integrates local spatial cues with gated state-space scans to improve multi-scale feature aggregation. Second, we introduce a Mask-Aware Height Refinement (MHR) module, which utilizes the segmentation output as a structure-aware confidence prior to clean regression artifacts. Finally, we provide a vastly expanded experimental evaluation, moving beyond the DFC23 \cite{dfc23} benchmark to include the DFC19 \cite{dfc19} and the challenging Huawei BHE datasets, offering a more rigorous analysis of the model's scalability and robustness.

Building on these advancements, the main contributions of this work are summarized as follows:
\begin{itemize}
    \item A unified RGB-only multi-task framework that explicitly couples segmentation and height regression through structural constraints, eliminating reliance on auxiliary modalities.
    \item A boundary-aware loss strategy that emphasizes height discontinuities along building edges, thereby reducing boundary smoothing and improving structural fidelity.
    \item The integration of a Mamba Attention Module (MAM) and a Spatial-Aware Mamba Feature Pyramid Network (S-MambaFPN) to enhance global--local feature interaction and multi-scale aggregation.
    \item A mask-aware height refinement module that utilizes segmentation outputs as structure-aware confidence priors for precise residual height correction.
    \item Extensive experimental validation across the DFC19, DFC23, and Huawei BHE benchmarks, establishing a new performance upper bound with IoU scores of 0.90, 0.93, 0.60, and RMSE of 1.06~m, 1.77~m and 9.23~m respectively, outperforming state-of-the-art baselines on the complex Huawei BHE dataset by 7.0\% in segmentation and 1.18~m in height estimation accuracy.
\end{itemize}

\section{Related Work}

\subsection{Height Estimation in Remote Sensing}

Estimating building height from monocular RGB satellite imagery is a fundamentally ill-posed problem, as three-dimensional structure must be inferred from appearance alone under a near top-down imaging geometry. Early approaches adapt convolutional encoder--decoder architectures to regress per-pixel height maps from overhead images \cite{IM2HEIGHT, IM2ELEVATION}. While these methods capture local texture and contextual cues, they remain highly sensitive to illumination changes, rooftop materials, and heterogeneous urban layouts, often producing unstable predictions and systematic underestimation of tall buildings.

To alleviate regression instability, several works reformulate height estimation using discretization or hybrid classification--regression strategies. Ordinal regression methods \cite{dorn} predict relative depth ordering rather than absolute values, while adaptive binning approaches such as AdaBins \cite{Adabins} and BinsFormer \cite{binsformer} dynamically allocate depth intervals to better handle scale variation. More recent hybrid frameworks \cite{chen2023htc} explicitly address the long-tailed distribution of building heights by separating foreground and background statistics. Although these techniques improve numerical stability, they remain appearance-driven and do not explicitly encode building structure, often struggling with boundary artifacts, tall-building sparsity, and generalization across cities and sensors.

Other lines of work incorporate auxiliary cues to reduce ambiguity. Shadow-based methods exploit geometric relationships between building height and cast shadows under known sun angles \cite{shadowbuilding, shadow}, while multi-modal approaches integrate SAR or LiDAR-derived information to inject explicit structural signals \cite{mftsc, tridentmtl}. However, such cues are frequently unavailable or unreliable at inference time, limiting their applicability in large-scale RGB-only scenarios.

Overall, despite architectural and formulation advances, monocular RGB height estimation remains weakly constrained by structure. Height is typically predicted independently of explicit building extents, leading to boundary bleeding, noisy regression near roof edges, and poor robustness under domain shifts. These limitations motivate approaches that more tightly couple height prediction with semantic building structure.

\subsection{Semantic Segmentation in Remote Sensing}

Semantic segmentation has seen significant progress in remote sensing, driven by architectures that combine fine-grained spatial modeling with global contextual reasoning. Classical segmentation pipelines typically rely on deep convolutional backbones such as ResNet \cite{resnet} combined with feature pyramid networks (FPN) for hierarchical multi-scale feature aggregation \cite{fpn}. These designs improve representation stability and enable the capture of spatial structures across multiple resolutions, which is essential for high-resolution satellite imagery.

More recently, transformer-based vision models  (Vision Transformers (ViT) \cite{vit} and Swin Transformer \cite{swin}) have been introduced to improve scene-level understanding by modeling global dependencies across large spatial extents. Building on these architectural advances, several remote sensing segmentation frameworks have been proposed to improve robustness in aerial imagery. Hybrid designs that integrate convolutional and transformer components, such as FarSeg and FarSeg++ \cite{FarSeg, FarSeg++} and ST-Unet \cite{ST-Unet}, demonstrate improved robustness under large-scale spatial variation and domain shifts. Additional aerial segmentation frameworks, including PointFlow \cite{PointFlow}, which refines segmentation through point-based feature propagation, and UANet \cite{uanet}, which incorporates uncertainty-aware decoding for building extraction, further improve robustness across diverse urban scenes.

While advances in segmentation significantly improve building footprint extraction, segmentation alone does not resolve height estimation ambiguity. However, accurate segmentation provides a strong structural prior: it defines valid spatial support for height prediction, sharpens object boundaries, and constrains height discontinuities to building extents. Prior work has shown that enforcing boundary-aware structural cues improves geometric consistency in building reconstruction \cite{boundaryaware}, while shadow-based approaches further exploit structural cues to infer building height from overhead imagery \cite{shadowbuilding}.

\begin{figure*}[ht]
  \centering
  \includegraphics[width=\linewidth]{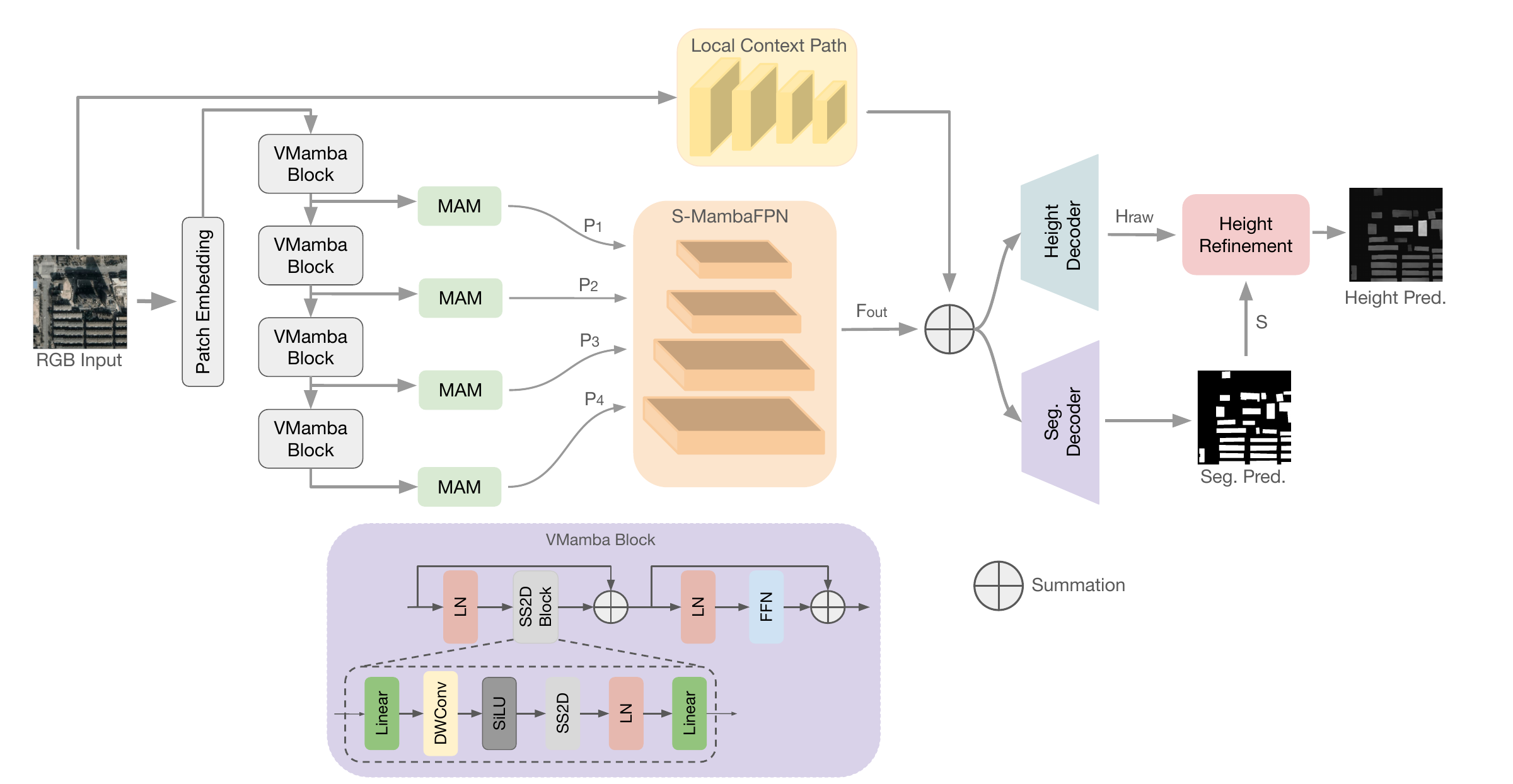}
  \caption{Overview of BuildMamba pipeline.}
  \label{fig:overview}
\end{figure*}

\subsection{Multi-Task Learning in Remote Sensing}

Given the strong structural dependency between building footprints and heights, multi-task learning frameworks have been proposed to jointly learn segmentation and height estimation. PopNet \cite{popnet} adopts an encoder--dual-decoder architecture in which semantic features guide height prediction through class-specific anchors. Other approaches introduce explicit structural constraints, such as boundary-aware decoders and gradient-based losses, to enforce geometric consistency along building edges \cite{boundaryaware}. LIGHT \cite{light} further explores gated cross-task interaction to enhance feature sharing between instance segmentation and height regression branches.

Several methods emphasize multi-scale and semantic guidance for height prediction. MFTSC \cite{mftsc} integrates hierarchical features and semantic priors to improve spatial consistency, while HGDNet \cite{hgdnet} introduces height-hierarchical segmentation as an auxiliary task to regularize regression under uneven height distributions. Contrastive learning has also been explored to align semantic and height-aware representations across scales \cite{jointcontrastive}.

Despite these advances, existing multi-task approaches often rely on shallow task coupling, handcrafted priors, or auxiliary supervision derived from DSMs or height categorization. Many lack robust global--local feature interaction and remain sensitive to domain shifts, particularly in RGB-only settings. Height estimation is frequently treated as an auxiliary output rather than a structurally constrained task, limiting the effectiveness of joint learning.

Building on these observations, the preliminary version of our framework \cite{buildmamba_conference}, demonstrated the feasibility of using a Visual State-Space backbone for joint building segmentation and height estimation. The current work significantly extends this direction by strengthening global--local feature aggregation and task coupling through a Spatial-Aware Mamba Feature Pyramid Network and a mask-aware height refinement strategy, enabling more robust and structurally consistent monocular height estimation from RGB satellite imagery.

In summary, prior work on building height estimation from remote sensing imagery highlights the inherent difficulty of monocular RGB inference and the limitations of appearance-driven regression, discretization-based formulations, and auxiliary cue exploitation. While advances in semantic segmentation and multi-task learning demonstrate that structural information can substantially improve height prediction, existing approaches often rely on shallow task coupling, handcrafted priors, or additional supervision that limits robustness and scalability in RGB-only settings. Moreover, many architectures struggle to simultaneously capture long-range contextual dependencies and fine-grained spatial detail, which is critical for resolving tall-building ambiguity and boundary artifacts in dense urban scenes. These observations motivate a unified architectural design that explicitly integrates global--local representation learning with strong structural coupling between segmentation and height estimation. In the following section, we introduce BuildMamba, a multi-task framework built upon a visual state-space backbone that addresses these challenges through efficient global modeling, spatially aware feature aggregation, and mask-guided height refinement.

\section{Methodology}
\begin{figure*}[ht]
  \centering
  \includegraphics[width=0.9\linewidth]{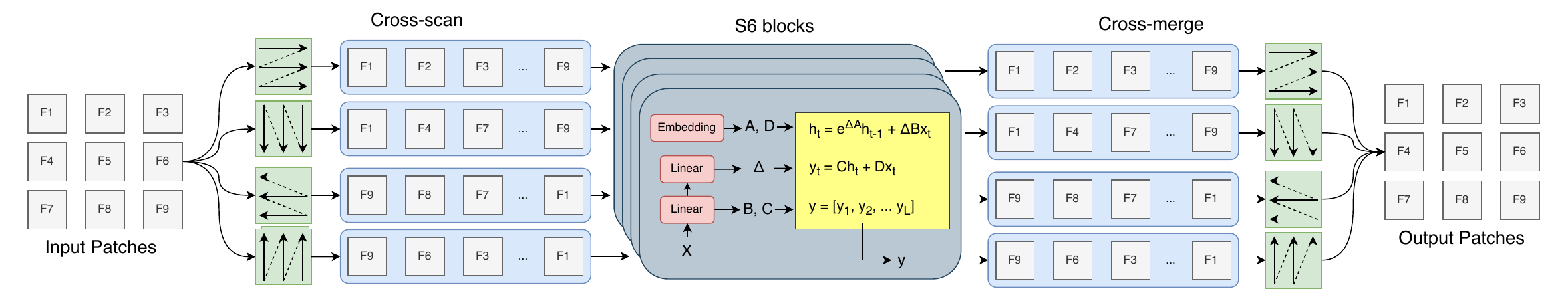}
  \caption{Illustration of 2D-Selective-Scan (SS2D) \cite{vmamba}.}
  \label{fig:ss2d}
\end{figure*}

This section outlines the proposed methodology, BuildMamba, which specifically targets the challenges of accurate height estimation from a single RGB image, and the need for stronger structural coupling in remote sensing imagery. As illustrated in Figure~\ref{fig:overview}, the model employs a dual-path encoder architecture that integrates a pre-trained visual state-space backbone (VMamba) to capture global long-range dependencies, alongside CNN-based modules to effectively model local feature context. Since VMamba does not employ an attention mechanism, we propose a spatial-based self-attention module, termed the Mamba Attention Module (MAM), to enhance spatial feature refinement. The refined feature maps from the VMamba backbone are further processed through a Mamba-based FPN to capture rich multi-scale contextual information. Following this, separate decoders are applied for building height estimation and semantic segmentation. Additionally, local feature maps extracted via the CNN branch are fused into both decoder pathways to improve spatial precision. To refine height maps, we employ a mask-aware refinement module that uses semantic segmentation maps as structural priors. This approach effectively eliminates high-frequency speckle noise in low-texture regions and sharpens building boundaries in the estimated height maps.

\subsection{Visual Mamba (VMamba)}

\textbf{Selective State-Space Models.} At the core of VMamba lies the Selective State-Space Model (S6), a computational mechanism that generalizes classical state-space models (SSMs) into a deep learning context. While traditional SSMs are linear time-invariant (LTI) systems, VMamba introduces an input-dependent selection mechanism, allowing the model to adaptively propagate information based on the current context.

Formally, the continuous-time state-space model maps a 1-D input function or sequence \( x(t) \in \mathbb{R} \) to an output \( y(t) \in \mathbb{R} \) through a latent hidden state \( h(t) \in \mathbb{R}^N \). This process is governed by the following ordinary differential equation (ODE) and a linear output projection:
\begin{equation} \label{ssm_continuous}
    \begin{aligned}
        h'(t) &= \mathbf{A}h(t) + \mathbf{B}x(t) \\
        y(t) &= \mathbf{C}h(t) + \mathbf{D}x(t)
    \end{aligned}
\end{equation}
In this formulation:
\begin{itemize}
    \item \( \mathbf{A} \in \mathbb{R}^{N \times N} \) is the state transition matrix (evolution parameter),
    \item \( \mathbf{B} \in \mathbb{R}^{N \times 1} \) is the input projection matrix,
    \item \( \mathbf{C} \in \mathbb{R}^{1 \times N} \) is the output projection matrix,
    \item \( \mathbf{D} \in \mathbb{R}^{1 \times 1} \) is a skip connection parameter.
\end{itemize}

To integrate this continuous system into a discrete deep learning framework, the ODE is discretized using a timescale parameter \( \Delta \), as depicted inside the S6 blocks of Figure~\ref{fig:ss2d}. Applying the Zero-Order Hold (ZOH) method, the continuous parameters are transformed into their discrete counterparts, yielding the discrete recurrence relation used in the forward pass:
\begin{equation} \label{ssm_discrete}
    h_t = \overline{\mathbf{A}} h_{t-1} + \overline{\mathbf{B}} x_t, \quad y_t = \mathbf{C} h_t + \mathbf{D} x_t,
\end{equation}
where $\overline{\mathbf{A}}$ and $\overline{\mathbf{B}}$ are obtained via discretization with time-scale $\Delta$.
Following \cite{vmamba}, we use
\begin{equation}\label{eq:ssm_discretization_vmamba}
\overline{\mathbf{A}} = e^{\mathbf{A}\Delta}, 
\qquad
\overline{\mathbf{B}} = \mathbf{B}\Delta,
\end{equation}
where $\overline{\mathbf{A}}$ is the exact ZOH discretization of $\mathbf{A}$ and $\overline{\mathbf{B}}=\mathbf{B}\Delta$ corresponds to the first-order Taylor approximation of the ZOH-discretized input term.

While the S6 mechanism effectively models 1D sequences, standard SSMs are inherently causal and cannot naturally accommodate the non-causal, 2D spatial relationships typical of visual data. To resolve this, VMamba employs the Cross-Scan Module (CSM) illustrated in Figure~\ref{fig:ss2d}. In this process, input image patches are unfolded into four distinct sequences via a Cross-Scan operation, traversing the 2D feature map along four different scanning paths. Each sequence is processed in parallel by a separate S6 block, enabling the model to capture global spatial dependencies from all directions simultaneously. Finally, a Cross-Merge operation reshapes and combines the resulting sequences to reconstruct the output 2D feature map.

\textbf{VMamba.} Visual Mamba (VMamba) \cite{vmamba} is a novel vision architecture designed to overcome the computational bottlenecks of traditional global modeling approaches. While Convolutional Neural Networks (CNNs) operate with efficient linear complexity, they are fundamentally limited by local receptive fields, which restricts their ability to capture long-range dependencies. Conversely, Vision Transformers (ViTs) \cite{vit} excel at modeling global context via self-attention mechanisms but suffer from quadratic computational complexity (\( O(N^2) \)) with respect to image resolution, creating a significant bottleneck for high-dimensional inputs. VMamba resolves this trade-off by integrating Selective State-Space Models (SSMs) into a hierarchical visual backbone, effectively combining the global receptive field of transformers with the linear computational efficiency (\( O(N) \)) of CNNs.

As illustrated in Figure~\ref{fig:overview}, the architecture is constructed to facilitate fine-to-coarse feature learning. The process begins with a patch embedding layer, which partitions the input image into non-overlapping patches and projects them into a lower-dimensional latent space. These tokenized representations are then processed through four distinct stages of Mamba blocks, each generating feature maps at progressively lower resolutions. This hierarchical structure produces multi-scale features that capture both fine-grained local details and high-level semantic information.

The core processing unit of these stages is the VMamba Block, which utilizes a residual architecture comprising a Selective Scan 2D (SS2D) module for spatial modeling and a Feed-Forward Network (FFN) for channel mixing. Layer Normalization (LN) is applied before each module to ensure training stability. The internal structure of the SS2D module adopts a streamlined design to control information flow: the input features undergo a linear projection, a Depthwise Convolution (DWConv) for local context, and a SiLU activation before passing through the core SS2D operation of Figure \ref{fig:ss2d}. The output of the scanning mechanism goes through additional LN and linear projection layers.

In this study, this VMamba \cite{vmamba} encoder, pre-trained on ImageNet, serves as the backbone for our multi-task learning network. The multi-scale features extracted from the four stages are fed into parallel decoders to simultaneously perform semantic segmentation and height regression. By leveraging the global receptive field of the SSM mechanism, VMamba \cite{vmamba}  significantly enhances the model's ability to resolve complex spatial dependencies inherent in satellite imagery, offering a robust balance between accuracy and computational speed.

\begin{figure*}[ht]
\centering
\resizebox{0.9\textwidth}{!}{%
\begin{tikzpicture}[
    node distance=10mm and 15mm,
    every node/.style={font=\small\sffamily},
    tensor/.style={draw, fill=#1, minimum width=1.5cm, minimum height=1.1cm, outer sep=0pt},
    op/.style={draw, circle, inner sep=0pt, minimum size=6mm},
    layer/.style={draw, fill=purple!20, minimum width=2.4cm, minimum height=0.8cm, align=center},
vpool/.style={
    draw,
    fill=green!15,
    minimum width=2.2cm,
    minimum height=0.45cm,
    align=center
},
hpool/.style={
    draw,
    fill=green!15,
    minimum width=0.45cm,
    minimum height=2.2cm,
    align=center
},
    softmax/.style={draw, fill=gray!30, minimum width=1.4cm, minimum height=0.6cm},
    arrow/.style={-{Latex[length=2mm]}, thick}
]

\node[tensor=red!20] (fin) at (0,0) {\large $F_{in}$};

\draw (fin.north west) -- ++(0.2,0.2)
      -- ($(fin.north east)+(0.2,0.2)$)[fill=red!10]
      -- (fin.north east);

\draw ($(fin.north east)+(0.2,0.2)$)
      -- ($(fin.south east)+(0.2,0.2)$)
      -- (fin.south east);

\node[left=2pt] at (fin.west) {C};
\node[above right=3.5pt and 1pt] at (fin.north) {W};
\node[above right=0.2pt and 3.5pt] at (fin.east) {H};

\coordinate (branch) at ($(fin.east) + (0.8, 0)$);
\draw[thick] (fin.east) -- (branch);


\node[
    vpool,
    scale=0.72,
    above right=6mm and 10mm of branch
] (poolH) {};

\node[above=5mm of poolH, font=\small] {Pooling};
\node[above=2.5mm of poolH, font=\scriptsize] {$B \times C \times H \times 1$};

\draw (poolH.north west) -- ++(0.15,0.15)
    -- ($(poolH.north east)+(0.15,0.15)$)[fill=green!10]
    -- (poolH.north east);

\draw ($(poolH.north east)+(0.15,0.15)$)
    -- ($(poolH.south east)+(0.15,0.15)$)
    -- (poolH.south east);

\node[layer, right=1.2cm of poolH] (fcH) {\scriptsize Fully Connected};
\node[above=4pt] at ($(poolH)!0.5!(fcH)$) {$W_H$};


\node[
    hpool,
    scale=0.72,
    below right=8mm and 16mm of branch
] (poolW) {};

\node[below=5mm of poolW, font=\small] {Pooling};
\node[below=2.5mm of poolW, font=\scriptsize] {$B \times C \times 1 \times W$};

\draw (poolW.north west) -- ++(0.15,0.15)
    -- ($(poolW.north east)+(0.15,0.15)$)[fill=green!10]
    -- (poolW.north east);

\draw ($(poolW.north east)+(0.15,0.15)$)
    -- ($(poolW.south east)+(0.15,0.15)$)
    -- (poolW.south east);

\node[layer, right=1.85cm of poolW] (fcW) {\scriptsize Fully Connected};
\node[above=4pt] at ($(poolW)!0.5!(fcW)$) {$W_W$};

\node[op, right=7.2cm of branch] (mult) {\Large $\otimes$};
\node[below right=1pt and 2pt of mult, font=\scriptsize, align=left] {$B \times C \times W \times H$};

\node[softmax, right=2.2cm of mult] (soft) {Softmax};

\node[op, right=1.2cm of soft] (alphamult) {\Large $\odot$};

\node[above=0.8cm of alphamult] (alpha) {$\alpha$};
\node[above=0.1cm of alpha, align=center, font=\scriptsize\bfseries]
{Learnable \\ Parameter};

\node[op, right=1.5cm of alphamult] (add) {\Large $\oplus$};


\node[tensor=orange!80!black] (fout) at (19.0,0) {\large $F_{out}$};

\draw (fout.north west) -- ++(0.2,0.2)
      -- ($(fout.north east)+(0.2,0.2)$)[fill=orange!40]
      -- (fout.north east);

\draw ($(fout.north east)+(0.2,0.2)$)
      -- ($(fout.south east)+(0.2,0.2)$)
      -- (fout.south east);

\node[above=2pt] at (fout.north) {W};
\node[above left=1pt and 2pt] at (fout.west) {C};
\node[above right=0.2pt and 3.5pt] at (fout.east) {H};


\draw[arrow] (branch) |- (poolH);
\draw[arrow] (branch) |- (poolW);

\draw[arrow] (poolH) -- (fcH);
\draw[arrow] (poolW) -- (fcW);

\draw[arrow] (fcH) -| node[above, font=\scriptsize] {$B \times H \times C$} (mult);
\draw[arrow] (fcW) -| node[below, font=\scriptsize] {$B \times W \times C$} (mult);

\draw[arrow] (mult) -- (soft);
\draw[arrow] (soft) -- (alphamult);

\draw[arrow] (alpha) -- (alphamult);

\draw[arrow] (fin.south) -- ++(0,-3.2) -| (alphamult.south);

\draw[arrow] (alphamult) -- (add);

\draw[arrow] (fin.north) -- ++(0,2.0) -| (add.north);

\draw[arrow] (add) -- (fout.west);

\end{tikzpicture}
}

\caption{Overview of the Mamba Attention Module (MAM).}
\label{fig:mam}
\end{figure*}
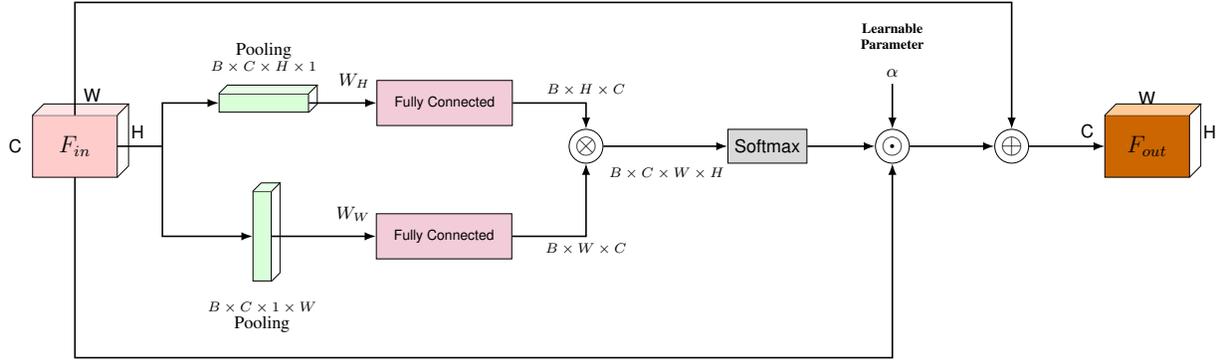

\subsection{Mamba Attention Module (MAM)}
To enhance the fusion of local and global features extracted by the dual-path encoder, a Mamba Attention Module (MAM) is integrated at the output of each stage in the VMamba encoder. Drawing inspiration from spatial attention mechanisms such as SAM \cite{sam}, MAM is designed to refine feature representations by dynamically recalibrating spatial importance, allowing the network to emphasize informative regions while suppressing less relevant or noisy features.

As illustrated in Figure~\ref{fig:mam}, MAM processes features along two pathways by applying adaptive average pooling separately across height (\( H \times 1 \)) and width (\( 1 \times W \)) dimensions. This dual-pathway approach captures spatial dependencies from both horizontal and vertical perspectives. The pooled features are transformed through fully connected layers and combined to generate an attention map, which highlights significant spatial regions. The final output is produced by applying this attention map to the input features along with a residual connection, effectively enhancing spatially relevant information. The attention map is normalized with a softmax function and subsequently applied to the original input features: 
\begin{equation}
\mathbf{F}_{\text{out}} =
\alpha \cdot 
\text{softmax}\big(\mathbf{W}_w\mathbf{P}_w\mathbf{F}_{\text{in}} \otimes \mathbf{W}_h\mathbf{P}_h\mathbf{F}_{\text{in}}\big) 
\odot \mathbf{F}_{\text{in}} 
+ \mathbf{F}_{\text{in}},
\end{equation}
where \(\mathbf{F}_{\text{out}} \) is the refined output feature map, and \( \mathbf{F}_{\text{in}} \) is the input feature map. \( \alpha \) is a learnable scalar parameter that controls the contribution of the attention mechanism. \( \mathbf{P}_w \) and \( \mathbf{P}_h \) are the adaptive average pooling operators along the height and width dimensions, respectively. \( \mathbf{W}_w \) and \( \mathbf{W}_h \) are learnable weight matrices applied via fully connected layers, responsible for transforming the pooled feature maps to capture richer dependencies before computing the attention map. The symbol \( \odot \) denotes element-wise multiplication, while \( \otimes \) represents the matrix multiplication.

\begin{figure*}[ht]
  \centering
  \includegraphics[width=0.8\linewidth]{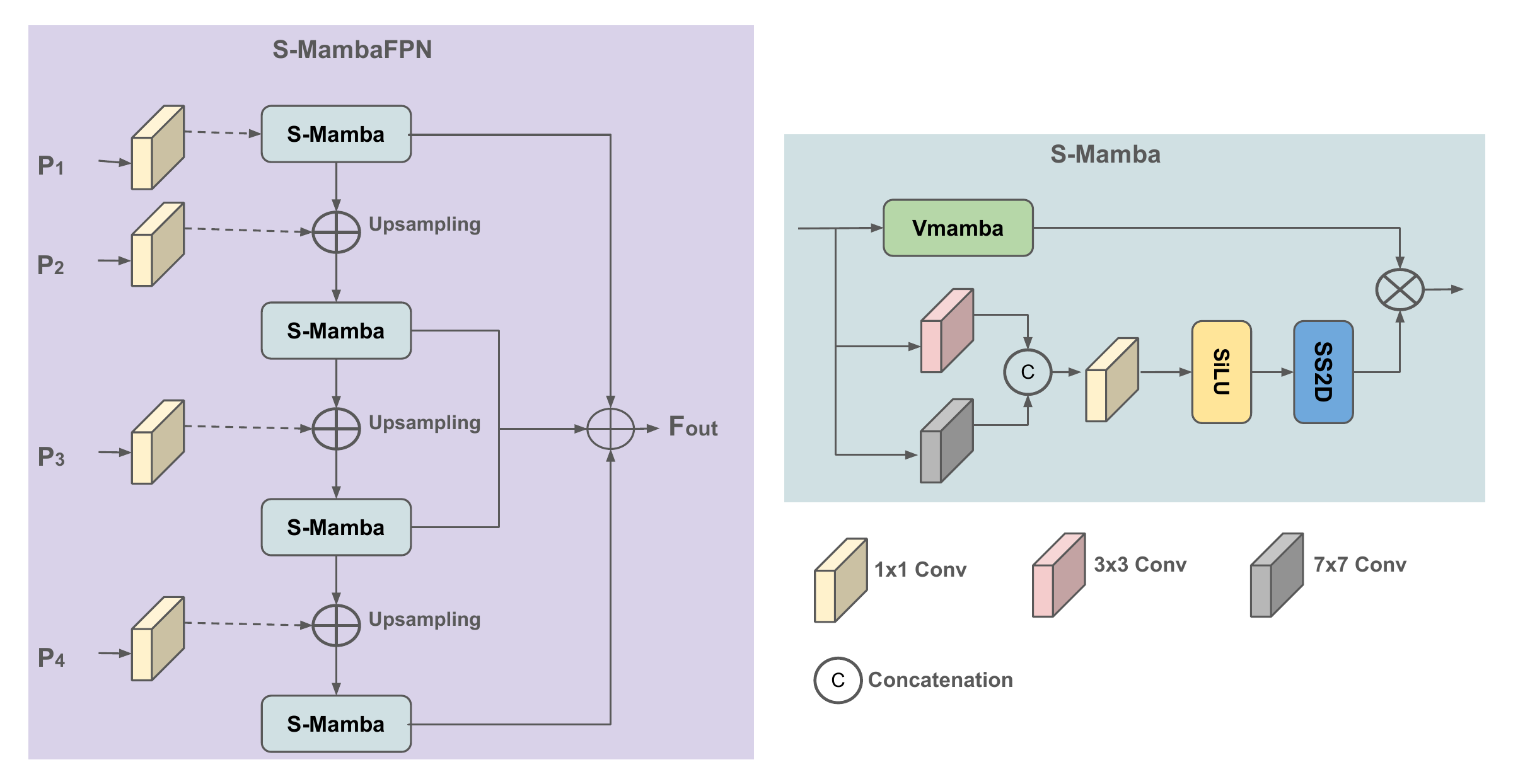}
  \caption{Overview of the S-MambaFPN module.}
  \label{fig:smambafpn}
\end{figure*}

\subsection{Spatial-Aware Mamba-FPN (S-MambaFPN)}
Multi-scale feature extraction is essential for dense prediction because objects, parts, and scene context span a wide range of sizes and spatial frequencies: high-resolution features preserve fine edges and small instances, while coarser resolutions capture category-level semantics and global context more reliably. Using a single-resolution representation causes scale mismatch—small targets get over-smoothed, large structures become fragmented—and increases aliasing when features are naively up/downsampled. To address this, we adopt a modified FPN  module to aggregate information across resolutions and maintain coherent semantics throughout the hierarchy, and we employ a shared FPN module for both the semantic segmentation and the height estimation branches to ensure consistent multi-scale conditioning and shared context.

Although FPN \cite{fpn} provides robust multi-scale routing, its refinement stage is inherently local and therefore limited in propagating long-range dependencies, particularly at coarser pyramid levels. To overcome this without changing the pyramid topology, we replace the refinement operator with the official VMamba \cite{vmamba} block and treat this configuration as our baseline Mamba-FPN. The state-space scanning mechanism in Vision Mamba efficiently spreads information over large spatial extents with linear complexity, endowing each FPN stage with long-range modeling while preserving the original fusion behavior. However, relying solely on this scan can under-capture isotropic local context and mid-range texture that remain crucial for precise localization and boundary fidelity.

To compensate for this locality gap, we introduce a spatial-aware scanning branch that complements the baseline Mamba branch (See Figure \ref{fig:smambafpn}). We first extract local and mid-range cues using depthwise $3\times3$ and $7\times7$ convolutions, concatenate the resulting maps, fuse them with a lightweight convolution followed by a SiLU activation, and then apply the SS2D scanning. The outputs of the spatial-aware branch and the baseline scan branch are combined via element-wise multiplication to gate long-range responses with spatial evidence, and stability is maintained by using the standard VMamba residual path, featuring pre-norm and stochastic depth. This output replaces the refinement operator at each pyramid level; the paired $3\times3/7\times7$ footprints align with heterogeneous pyramid strides—capturing fine boundaries at higher resolutions and stabilizing broader context at coarser scales—so spatial cues and long-range dependencies are fused within a single FPN  stage.

\begin{figure*}[ht]
  \centering
  \includegraphics[width=0.6\linewidth]{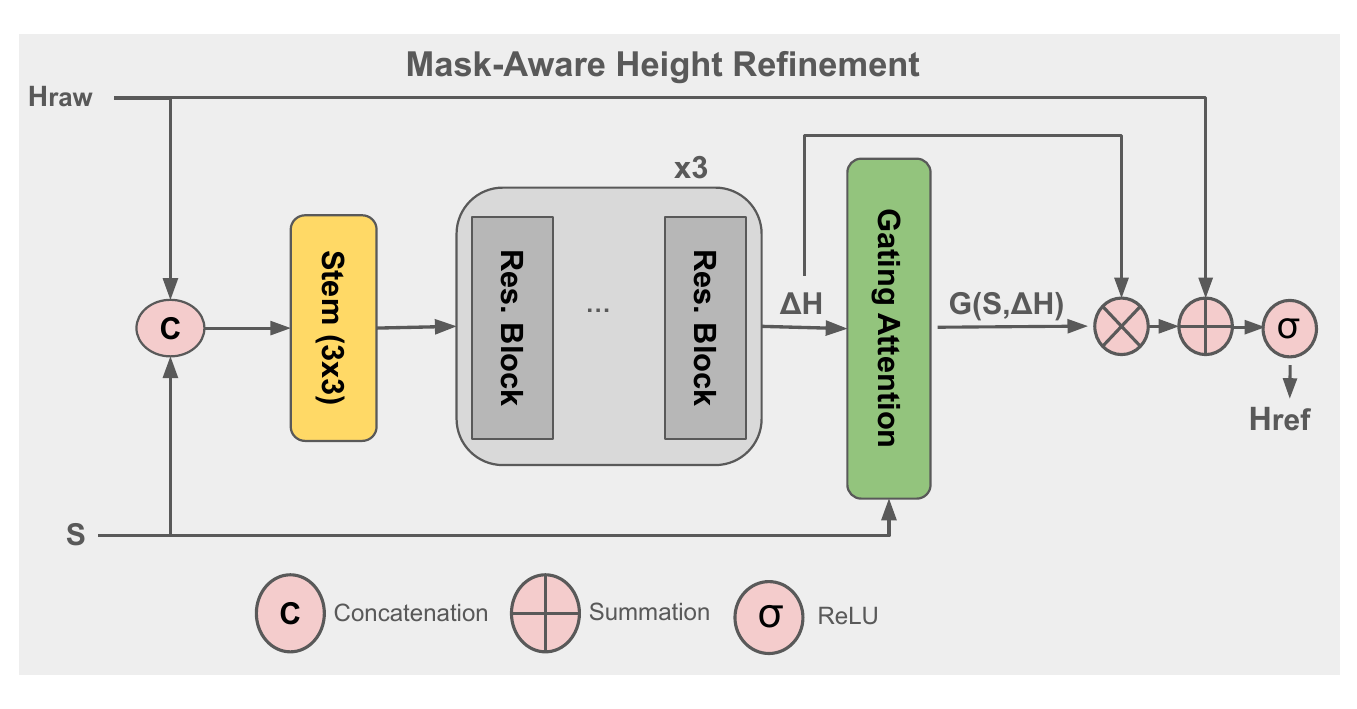}
  \caption{Overview of the Mask-Aware Height Refinement (MHR) module.}
  \label{fig:refinement}
\end{figure*}

\subsection{Mask-Aware Height Refinement}

Height estimation in our multi-task setting exhibits characteristic failure modes: high-frequency speckle in low-texture areas and boundary bleeding across class edges, whereas the semantic mask converges faster and encodes object supports and boundaries more reliably. We exploit this asymmetry with a mask-aware refinement module, shown in Figure \ref{fig:refinement}, that lets the height branch communicate with the segmentation output, but without collapsing height estimation into a trivial function of the mask. Concretely, the segmentation mask acts as a structure-aware prior that steers residual corrections, while the refiner remains height-centric: it operates on $(H_{\text{raw}}, S)$ but is supervised purely on height, so it learns to clean the height map instead of re-implementing segmentation. This design also allows the refiner to actively suppress false-positive segmentation responses on the height side, e.g., dampening spurious elevated predictions where the mask incorrectly fires.

Let $H_{\text{raw}}\in\mathbb{R}^{1\times H\times W}$ denote the raw height prediction and $S\in[0,1]^{1\times H\times W}$ the soft segmentation mask. We treat $S$ as a confidence map and construct a two-channel input $[H_{\text{raw}}, S]$, which is processed by a lightweight refinement network: a $3\times3$ stem of modest width followed by two residual blocks, optionally with dilations to capture meso-scale context while retaining locality. This prediction head outputs a residual:
\begin{equation}
    \Delta H \;=\; f_{\theta}\big([H_{\text{raw}}, S]\big).
\end{equation}

To couple correction strength to segmentation confidence while keeping the module height-focused, this spatial height residual $\Delta H$  is then modulated by a gated update function:
\begin{equation}
    G(S, \Delta H) = \left( \varepsilon + (1-\varepsilon) S^{\gamma} \right)  \Delta H,
\end{equation}
where $S$ represents the segmentation confidence, with $\varepsilon\in(0,1)$ providing a conservative floor and $\gamma>0$ controlling how sharply high-confidence regions receive stronger updates. This mask-aware gate is conceptually related to the confidence-weighted gating attention used in TransAdapter for feature-centric domain alignment~\cite{doruk2024transadapter}, but here it operates in the spatial domain by explicitly fusing $H_{raw}$ and $S$ to modulate height residuals rather than transformer features. The refined height is then obtained via a constrained residual update:
\begin{equation}
    H_{\text{ref}}\;=\;\operatorname{ReLU}\big(H_{\text{raw}} \;+\; G(S, \Delta H\big)\,\Delta H\big),
\end{equation}
which enforces non-negativity for absolute heights (or can be replaced by a range clamp when signed heights are used).

This sequence—use $S$ as a normalized confidence prior, concatenate it with $H_{\text{raw}}$, predict a residual through a shallow residual stack, gate that residual by segmentation confidence, and apply a constrained residual addition—keeps parameters and FLOPs minimal while ensuring: (i) height-centric behavior, since the module learns residuals on top of $H_{\text{raw}}$ rather than regressing heights directly from the mask; (ii) mask-aware corrections, concentrating strong updates on reliable object supports and boundaries; and (iii) false-positive suppression, as the joint dependence on $H_{\text{raw}}$ and $S$ allows the network to reduce heights where the mask is active but the height evidence is weak. In practice, this yields visibly cleaner height maps—reduced speckle on uniform surfaces, sharper discontinuities at class borders, and dampened artifacts in regions dominated by mask errors—while consistently improving quantitative height-estimation metrics without destabilizing training or incurring significant computational overhead.

\subsection{Objective Function}

To effectively train the BuildMamba framework, we formulate a multi-task objective function that jointly optimizes for semantic segmentation accuracy and height estimation precision. This composite loss, $\mathcal{L}_{\text{total}}$, is designed to address the specific challenges, including severe class imbalance, structural ambiguity at building boundaries, and the long-tailed nature of urban height distributions. The total objective is  summation of the segmentation and regression losses, formulated as: \begin{equation}
\mathcal{L}_{\text{total}} = \mathcal{L}_{\text{seg}} + \mathcal{L}_{\text{reg}}.
\end{equation}
\noindent \textbf{Semantic Segmentation Loss:} To ensure robust pixel-wise classification and precise boundary delineation, we employ a hybrid loss strategy. We combine the  Cross-Entropy loss ($\mathcal{L}_{\text{ce}}$) to supervise pixel probabilities and the Dice loss ($\mathcal{L}_{\text{dice}}$) to maximize the Intersection-over-Union (IoU) on imbalanced data. Furthermore, to enforce structural consistency, we introduce a boundary-aware term. A Laplacian kernel $\mathbf{L}$ is applied to both the ground truth  and predicted masks to extract second-order derivative edge maps; Binary Cross-Entropy loss ($\mathcal{L}_{\text{edge}}$) is then computed on these edge maps to sharpen building contours. The segmentation loss is defined as:

\begin{eqnarray}
\mathcal{L}_{\text{ce}} &\!\!\! = \!\!\!& -\sum_{i} \left[ S_i \log(\hat{S}_i) + (1-S_i) \log(1-\hat{S}_i)\right], \label{eq:ce} \\
\mathcal{L}_{\text{dice}} & \!\!\!= \!\!\!& 1 - \frac{2 \sum_i S_i \hat{S}_i}{\sum_i S_i + \sum_i \hat{S}_i}, \label{eq:dice} \\
\mathcal{L}_{\text{edge}} &\!\!\! =\!\!\! & -\sum_i \left[ \ell_i \log(\hat{\ell}_i) + (1 - \ell_i) \log(1 - \hat{\ell}_i) \right], \label{eq:bce} \\
\mathcal{L}_{\text{seg}} & \!\!\!=\!\!\! & \mathcal{L}_{\text{ce}} + \mathcal{L}_{\text{dice}} + 10 \mathcal{L}_{\text{edge}}, \label{eq:seg_total}
\end{eqnarray}
\noindent{where:}

\begin{itemize}
    \item $S_i$: Ground truth building label for pixel $i$.
    \item $\hat{S}_i$: Predicted probability of being building for pixel $i$.
    \item $\ell_i  =  \mathbf{L}(S)_i$: Laplacian edge value of ground truth at $i$.
    \item $\hat{\ell}_i = \mathbf{L}(\hat{S})_i$: Laplacian edge value of prediction at  $i$.
  
\end{itemize}

\noindent \textbf{Height Regression Loss:} For the height estimation task, we address the challenge of outliers and long-tailed height distributions. Instead of the standard Mean Squared Error, we utilize the Huber loss ($\mathcal{L}_{\text{reg}}$), which behaves as a quadratic loss for small errors and a linear loss for large errors. This formulation ensures differentiability while preventing extreme height values from destabilizing the gradient descent. The regression loss is calculated as:

\begin{equation}
\mathcal{L}_{\text{reg}}(H, \hat{H}) = \frac{1}{N} \sum_{i} 
\begin{cases} 
\frac{1}{2} (H_i - \hat{H}_i)^2, & \text{if } |H_i - \hat{H}_i| < \delta \\
\delta |H_i - \hat{H}_i| - \frac{1}{2} \delta^2, & \text{otherwise}
\end{cases}
\label{eq:huber}
\end{equation}

\noindent where $H$ and $\hat{H}$ represent the ground truth and predicted height maps, and $\delta$ controls the transition between the quadratic and linear phases.
\section{Simulation Results and Discussions}

\subsection{Datasets}

\subsubsection{DFC19}
The DFC19 (Data Fusion Contest 2019) dataset \cite{dfc19} is a large-scale remote sensing benchmark that provides multi-view satellite imagery along with ground truth geometric and semantic annotations for urban regions in Jacksonville, Florida, and Omaha, Nebraska. The dataset spans approximately 100 km² and includes imagery collected between 2014 and 2016 at a ground sampling distance (GSD) of 1.3 meters. It comprises 2,783 triplets of RGB images, normalized DSMs representing above-ground-level height annotations, and semantic maps that preserve detailed building footprints, all at a resolution of 1024×1024 pixels.

The DFC19 dataset originally includes multi-class semantic labels and nDSMs designed for general-purpose 3D reconstruction tasks. To tailor the dataset specifically for building footprint extraction and height estimation, we applied a preprocessing pipeline to isolate building labels from the semantic maps. To prepare the dataset for training, NaN values in the nDSMs are imputed using nearest neighbor interpolation to ensure completeness. A binarization step is then applied to the semantic maps to isolate building footprints, simplifying the task to pure building extraction and height estimation. For model training, a random cropping strategy is adopted, generating 512×512 pixel crops from the original tiles. The number of crops per image is computed dynamically to meet a target of 20,000 total crops, which are subsequently split into 14000 training, 2000 validation, and 4000 testing subsets, while making sure that each subset contains crops from distinct set of images.

\subsubsection{DFC23}
The DFC23 (Data Fusion Contest 2023) Track-2 dataset \cite{dfc23}, released by IEEE GRSS, provides multi-source remote sensing data aimed at building height estimation and instance segmentation tasks. While the original dataset includes synthetic aperture radar (SAR) imagery, this study focuses exclusively on using RGB satellite images to predict both height maps and semantic labels. For compatibility with the semantic segmentation task, the instance segmentation labels are converted into semantic masks. The training dataset is split into 1,417 training images, 178 validation images, and 178 test images, as reference labels for the official validation and test sets are unavailable.

\begin{table}[ht]
\centering
\caption{Huawei BHE train/val and test set statistics.}
\label{tab:dataset_stats_transposed}
\small
\begin{tabular}{@{}lccc@{}}
\toprule
\textbf{Metric} & \textbf{Train/Val} & \textbf{Test} \\ \midrule
Total Buildings          & 177,408 & 58,399          \\
Avg. Buildings    & 31.61   & 22.18            \\
Building Pixel Ratio     & 0.24            & 0.29          \\
Maximum Height (m)          & 309              & 328    \\
Mean Height (m)              & 14.85          & 18.69  \\
Std. Deviation (m)           & 17.55            & 20.56   \\ \bottomrule
\label{table:huaweidatasetstats}
\end{tabular}
\end{table}

\subsubsection{Huawei BHE}
\begin{figure}[h]
\vspace{-1cm}
  \centering
 \includegraphics[width=\linewidth]{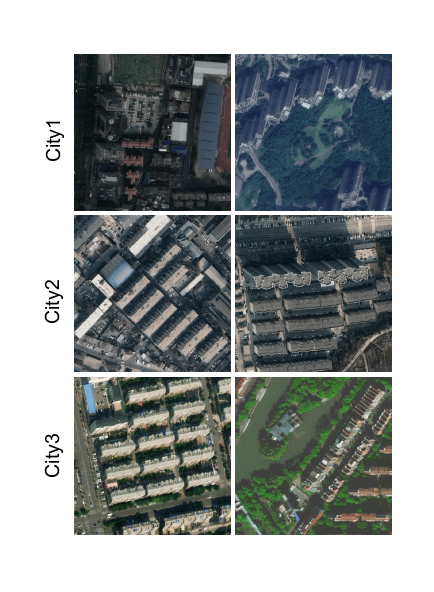}
   \vspace{-1.4cm}
  \caption{Sample images from different cities of  Huawei dataset.}
  \label{fig:samples}
\end{figure}

The Huawei BHE (Building Height Estimation) dataset comprises a comprehensive and diverse set of images collected from 14 different cities, carefully chosen to represent a wide range of urban environments. These cities exhibit substantial variation in sun elevation, illumination conditions, building height distributions, and structural orientations. While some locations contain predominantly nadir (top-down) views of buildings, others include oblique viewing angles, which further complicate the task of accurate building height estimation. All imagery is provided at level-18 resolution, offering fine spatial detail at a fixed size of 512×512. However, the annotations of the dataset present several inherent challenges that require robust modeling; specifically, it contains instances of mislabeling, inaccurate/missing height values, and temporal discrepancies where the ground truth labels and the satellite imagery originate from different time periods. For experimental evaluation, City1 is used as the validation set (1,975 images), City2 as the test set (2,633 images), and the rest as train set (3,637 images), allowing for a rigorous assessment of model generalization across heterogeneous urban and environmental conditions. Overall, the dataset serves as a challenging and valuable benchmark for urban structure analysis and height estimation in remote sensing. Simulation results highlight the robustness of BuildMamba model for such a varied and weakly labeled dataset. Figure \ref{fig:samples} shows sample images from different cities of Huawei BHE. 

Table \ref{table:huaweidatasetstats} reveals significant scale and structural variation between the training/validation and test sets. The training and validation sets, derived from a combination of different cities, comprise a total of 177,408 buildings with an average density of 31.61 buildings per image. In contrast, the test set presents a more challenging urban landscape for the model, characterized by higher vertical complexity. Specifically, the test set exhibits a greater maximum building height of 328~m compared to the train/val set's 309~m, alongside a higher mean height of 18.69~m and a broader standard deviation of 20.56~m. These metrics indicate that the model evaluation is performed on more diverse and varied urban morphologies than those predominantly encountered during the training phase.

\subsection{Implementation Details}
BuildMamba was separately trained and evaluated on the DFC19, DFC23, and Huawei BHE  datasets. The training setup was implemented on a NVIDIA GeForce RTX 2080 Ti. The initial learning rate was set to 5e-4 for the entire network, while a lower learning rate of 5e-5 was applied to the pre-trained VMamba backbone for fine-tuning. A weight decay of 0.0025 was employed to regularize the model and prevent overfitting. The training process was configured to run for a maximum of 465 epochs for DFC23, 225 epochs for DFC19, and 200 epochs for Huawei BHE, which were determined to be sufficient for convergence. The batch size was set to 8. A Cosine Annealing Warm Restarts scheduler was used. The scheduler restarted after 15 epochs, with the restart interval progressively doubling after each restart.

\subsection{Metrics}
Model performances are evaluated utilizing the standard metrics: IoU, F1 scores for semantic segmentation task, RMSE and delta scores for building height estimation. The mathematical definitions are as follows:
\begin{itemize}
    \item \textbf{ Intersection over Union (IoU):}
    \begin{equation}
        \text{IoU} = \frac{TP}{TP + FP + FN},
    \end{equation}
    where $TP$ (True Positives) represents correctly identified building pixels, $FP$ (False Positives) represents background pixels incorrectly classified as building, and $FN$ (False Negatives) represents building pixels that were missed.

    \item \textbf{F1 Score:}
    \begin{equation}
        \text{F1} = \frac{2 \cdot \text{Precision} \cdot \text{Recall}}{\text{Precision} + \text{Recall}} = \frac{2TP}{2TP + FP + FN},
    \end{equation}
    where \( TP \), \( FP \), and \( FN \) are the total number of true positives, false positives, and false negatives for the building class.

    \item \textbf{Root Mean Square Error (RMSE):}
    \begin{equation}
        \text{RMSE} = \sqrt{ \frac{1}{N} \sum_{i=1}^{N} (H_i - \hat{H}_i)^2 },
    \end{equation}
    where \( N \) is the total number of pixels per image, \( H_i \) is the ground truth height, and \( \hat{H}_i \) is the predicted height.
    \item \textbf{Delta Accuracy:}
    \begin{equation}
        \delta_1 = \frac{1}{N} \sum_{i=1}^{N} \mathbb{I} \left( \max\left( \frac{H_i}{\hat{H}_i}, \frac{\hat{H}_i}{H_i} \right) < 1.25 \right),
    \end{equation}
    \begin{equation}
        \delta_2 = \frac{1}{N} \sum_{i=1}^{N} \mathbb{I} \left( \max\left( \frac{H_i}{\hat{H}_i}, \frac{\hat{H}_i}{H_i} \right) < 1.25^2 \right),
    \end{equation}
    \begin{equation}
        \delta_3 = \frac{1}{N} \sum_{i=1}^{N} \mathbb{I} \left( \max\left( \frac{H_i}{\hat{H}_i}, \frac{\hat{H}_i}{H_i} \right) < 1.25^3 \right),
    \end{equation}
    where \( \mathbb{I}(\cdot) \) is the indicator function, which returns 1 if the condition is true and 0 otherwise. These metrics indicate the percentage of pixels where the predicted height is within a threshold factor of the ground truth.

\end{itemize}

\begin{table*}[t]
\centering
\caption{Semantic segmentation performance comparison across DFC19, DFC23, and Huawei BHE datasets. The best and second-best results are highlighted in bold and underlined, respectively. }
\label{tab:results_comparison}
\begin{tabular}{@{}lccccccc@{}} 
\toprule
& & \multicolumn{2}{c}{\textbf{DFC19}} & \multicolumn{2}{c}{\textbf{DFC23}} & \multicolumn{2}{c}{\textbf{Huawei BHE}} \\
\cmidrule(lr){3-4} \cmidrule(lr){5-6} \cmidrule(l){7-8} 
\textbf{Model} & \textbf{Input} & IoU $\uparrow$ & F1 $\uparrow$ & IoU $\uparrow$ & F1 $\uparrow$ & IoU $\uparrow$ & F1 $\uparrow$ \\
\midrule
BuildFormer   & RGB & 0.79 & 0.88 & 0.78 & 0.88 & 0.42 & 0.60 \\
DSMNet        & RGB & 0.45 & 0.62 & 0.85 & 0.92 & 0.42 & 0.60 \\
ST-UNet       & RGB & 0.84 & 0.89 & 0.82 & 0.90 & 0.52 & 0.68 \\
FarSeg++      & RGB & 0.82 & 0.90 & 0.83 & 0.91 & 0.52 & 0.68 \\
UANet         & RGB & \underline{0.89} & \underline{0.94} & \underline{0.92} & \textbf{0.96} & \underline{0.53} & \underline{0.69} \\
\midrule
\rowcolor{black!10}
\textbf{BuildMamba} & RGB & \textbf{0.90} & \textbf{0.95} & \textbf{0.93} & \textbf{0.96} & \textbf{0.60}  & \textbf{0.74} \\
\bottomrule
\end{tabular}
\end{table*}

\begin{table*}[t]
\centering
\caption{Height estimation performance comparison: RMSE (m) and Deltas ($\delta_n$). Among models with RGB-only input, the best and second-best results are highlighted in bold and underlined, respectively. }
\label{tab:height_results_comparison}
\setlength{\tabcolsep}{3.5pt} 
\begin{tabular}{@{}lccccccccccccc@{}} 
\toprule
& & \multicolumn{4}{c}{\textbf{DFC19}} & \multicolumn{4}{c}{\textbf{DFC23}} & \multicolumn{4}{c}{\textbf{Huawei BHE}} \\
\cmidrule(lr){3-6} \cmidrule(lr){7-10} \cmidrule(l){11-14} 
\textbf{Model} & \textbf{Input} & RMSE $\downarrow$ & $\delta_1 \uparrow$ & $\delta_2 \uparrow$ & $\delta_3 \uparrow$ & RMSE $\downarrow$ & $\delta_1 \uparrow$ & $\delta_2 \uparrow$ & $\delta_3 \uparrow$ & RMSE $\downarrow$ & $\delta_1 \uparrow$ & $\delta_2 \uparrow$ & $\delta_3 \uparrow$ \\
\midrule
LUMNet & RGB+Mask & {1.040} & {0.942} & {0.963} & {0.971} & 3.620 & {0.746} & 0.806 & 0.833 & {8.521} & {0.818} & {0.854} & {0.887} \\
\midrule
\midrule
DSMNet & RGB & 3.880 & 0.443 & 0.476 & 0.500 & 3.389 & {0.746} & 0.805 & 0.849 & \underline{10.413} & 0.609 & 0.641 & 0.671 \\
IM2HEIGHT & RGB & 4.899 & 0.201 & 0.203 & 0.205 & 7.210 & 0.678 & 0.679 & 0.681 & 12.301 & 0.385 & 0.385 & 0.385 \\
IM2ELEVATION & RGB & 4.952 & 0.774 & 0.774 & 0.774 & 6.501 & 0.759 & 0.764 & 0.772 & 11.841 & \underline{0.792} & \underline{0.798} & \underline{0.802} \\
HTC-DC & RGB & \underline{1.227} & \underline{0.934} & \underline{0.946} & \underline{0.950} & \underline{2.592} & \underline{0.827} & \underline{0.860} & \underline{0.873} &  10.621 & 0.479 & 0.512 & 0.544 \\
\midrule
\rowcolor{black!10}
\textbf{BuildMamba} & RGB & \textbf{1.058} & \textbf{0.939} & \textbf{0.965} & \textbf{0.968} & \textbf{1.772} & \textbf{0.881} & \textbf{0.902} & \textbf{0.932} & \textbf{9.230} & \textbf{0.806} & \textbf{0.824} & \textbf{0.860} \\
\bottomrule
\end{tabular}
\end{table*}

\subsection{Quantitative Results}
\textbf{Segmentation results.} To assess the efficacy of the proposed multi-task framework, we first evaluate the semantic segmentation performance of BuildMamba against five state-of-the-art baselines: BuildFormer \cite{BuildFormer}, DSMNet \cite{dsmnet}, ST-UNet \cite{ST-Unet}, FarSeg++ \cite{FarSeg++}, and UANet \cite{uanet}. As detailed in Table~\ref{tab:results_comparison}, evaluated across three heterogeneous datasets, BuildMamba consistently establishes a new performance upper bound, achieving peak IoU scores of 0.90 and 0.93 on the DFC19 and DFC23  benchmarks, respectively. While established models like UANet  and FarSeg++  yield competitive results on standard benchmarks, they exhibit limited robustness when transitioned to the more complex and weakly labeled Huawei BHE dataset that contains erroneous and missing building masks. In contrast, our model maintains high fidelity in building extraction, achieving an IoU of 0.60 and an F1-score of 0.74 on the Huawei BHE dataset, outperforming the second-best model UANet  by a significant margin of 7.0\% in IoU. These results suggest that the selective scanning mechanism inherent in the Mamba-based architecture effectively captures the intricate geometric features necessary for building delineation. Furthermore, the robust performance across both standard and challenging datasets highlights that our multi-task backbone effectively mitigates the performance trade-offs often encountered in joint segmentation and height prediction tasks, providing a more discriminative feature representation for the segmentation head.

\begin{figure*}[t]
\centering

\begin{subfigure}{0.13\textwidth}
\includegraphics[width=\textwidth] {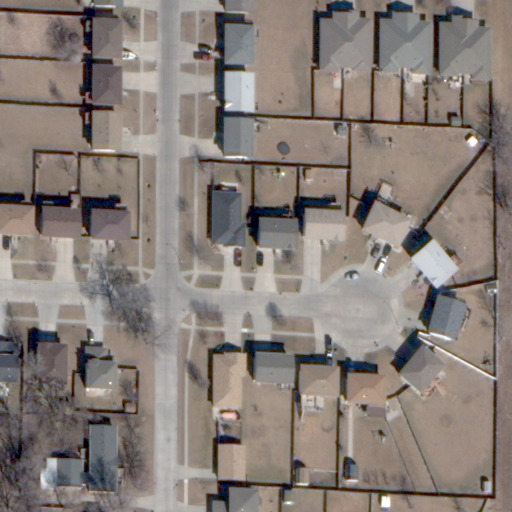}\caption*{RGB }\label{fig:rgb_seg}
\end{subfigure}
\begin{subfigure}{0.13\textwidth}
\includegraphics[width=\textwidth] {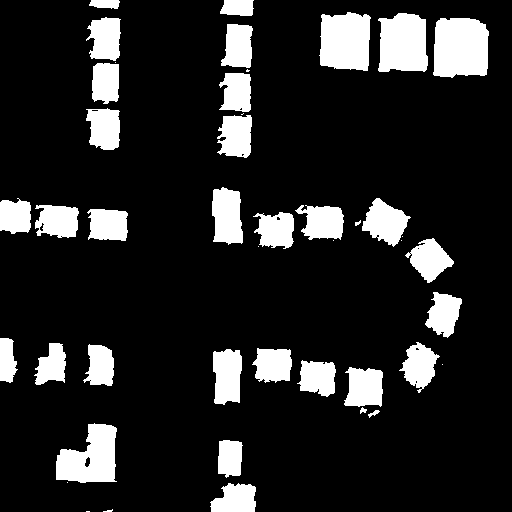}\caption*{Ground Truth}\label{fig:gt_seg}
\end{subfigure}
\begin{subfigure}{0.13\textwidth}
\includegraphics[width=\textwidth] {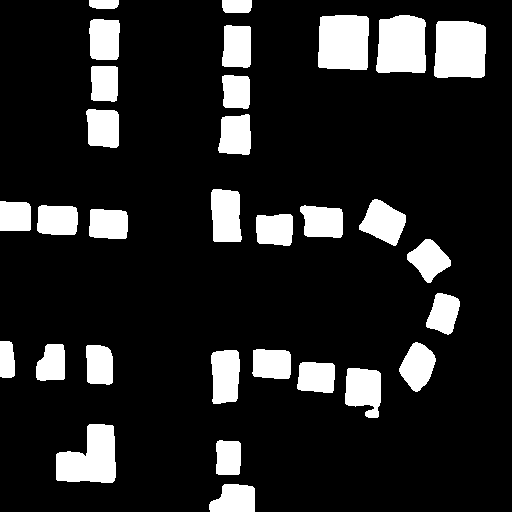}\caption*{UANet}\label{fig:uanet}
\end{subfigure}
\begin{subfigure}{0.13\textwidth}
\includegraphics[width=\textwidth] {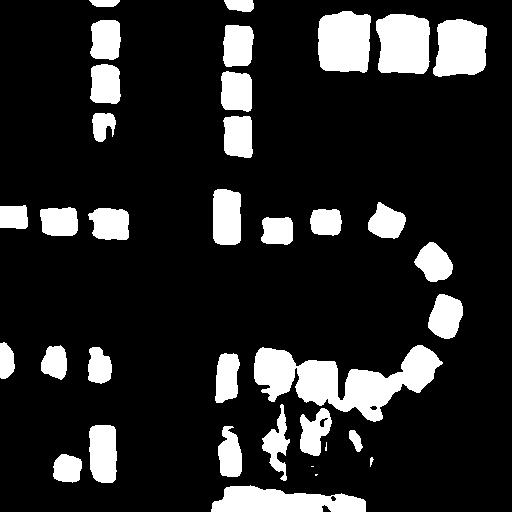}\caption*{BuildFormer}\label{fig:buildformer}
\end{subfigure}

\begin{subfigure}{0.13\textwidth}
\includegraphics[width=\textwidth] {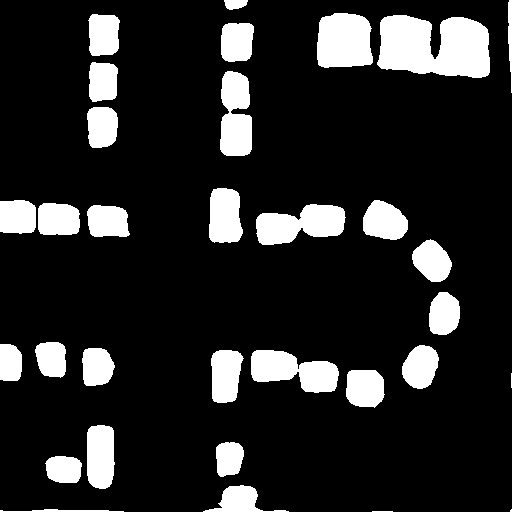}\caption*{FarSeg++}\label{fig:farsegpp}
\end{subfigure}
\begin{subfigure}{0.13\textwidth}
\includegraphics[width=\textwidth] {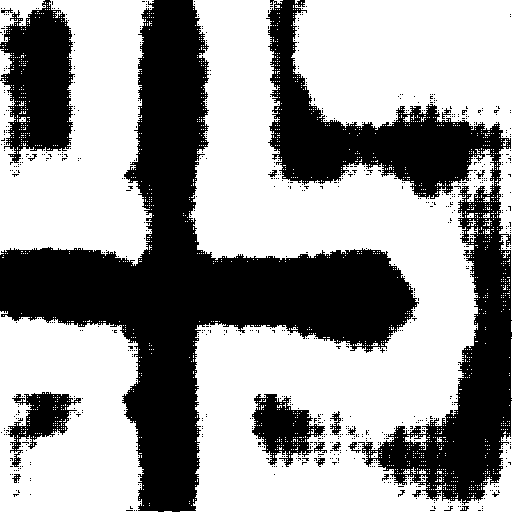}\caption*{DSMNet}\label{fig:dsmnet}
\end{subfigure}
\begin{subfigure}{0.13\textwidth}
\includegraphics[width=\textwidth] {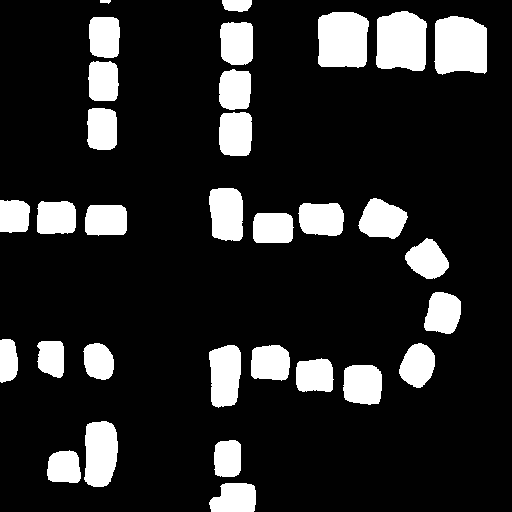}\caption*{ST-UNet}\label{fig:stunet}
\end{subfigure}
\begin{subfigure}{0.13\textwidth}
\includegraphics[width=\textwidth] {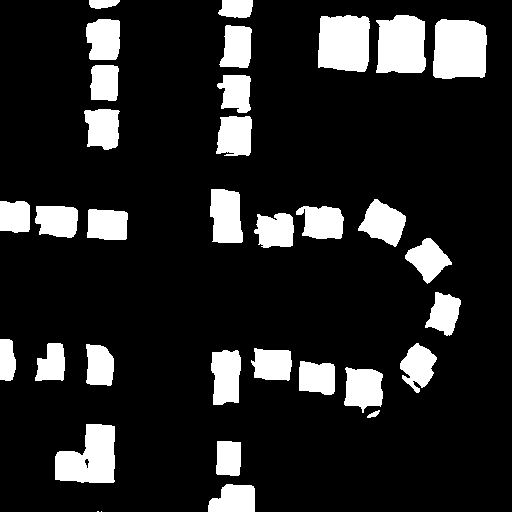}\caption*{\textbf{BuildMamba}}\label{fig:buildmamba}
\end{subfigure}

\caption{Visual comparison of segmentation results on DFC19 dataset.}
\label{fig:visualseg}
\end{figure*}

\begin{figure*}[h]
\centering

\begin{subfigure}{0.13\textwidth}
\includegraphics[width=\textwidth] {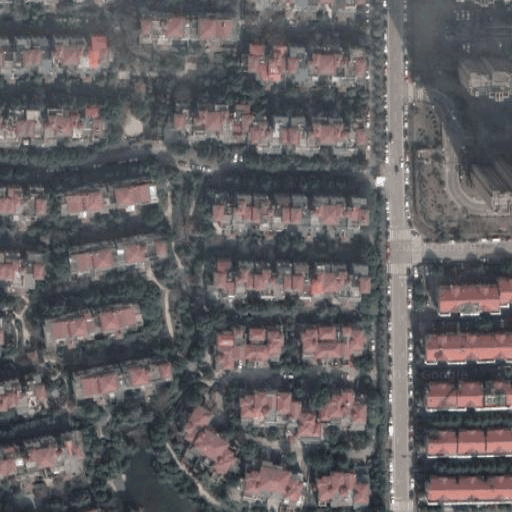}\caption*{RGB  }\label{fig:rgb_seg}
\end{subfigure}
\begin{subfigure}{0.13\textwidth}
\includegraphics[width=\textwidth] {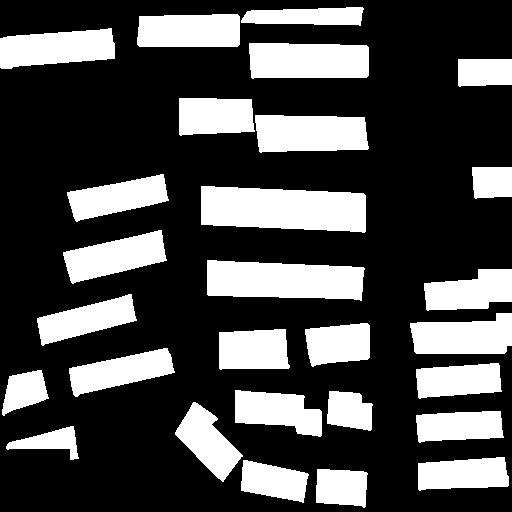}\caption*{Ground Truth}\label{fig:gt_seg}
\end{subfigure}
\begin{subfigure}{0.13\textwidth}
\includegraphics[width=\textwidth] {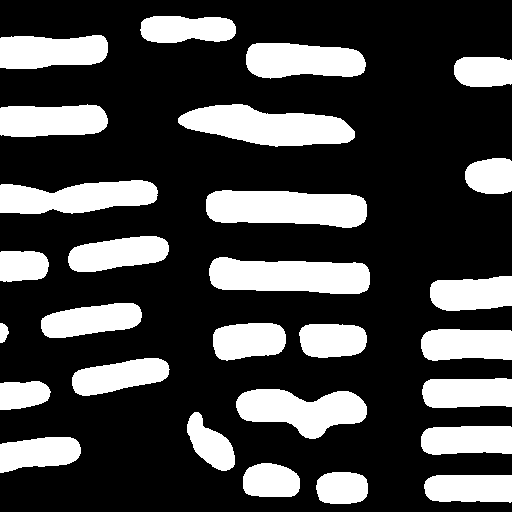}\caption*{UANet}\label{fig:uanet}
\end{subfigure}
\begin{subfigure}{0.13\textwidth}
\includegraphics[width=\textwidth] {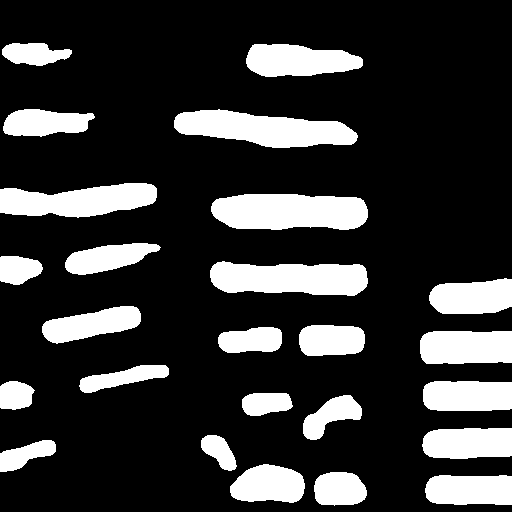}\caption*{BuildFormer}\label{fig:buildformer}
\end{subfigure}

\begin{subfigure}{0.13\textwidth}
\includegraphics[width=\textwidth] {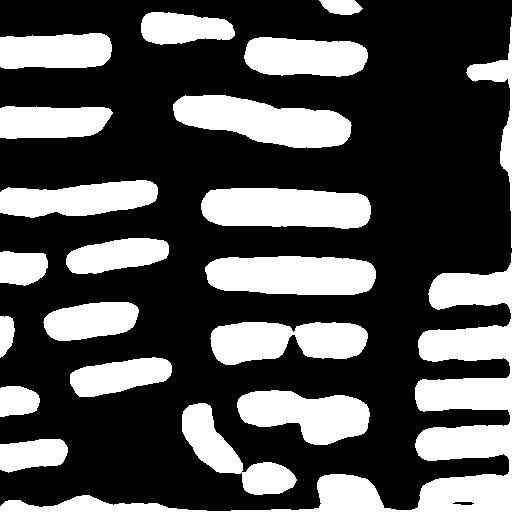}\caption*{FarSeg++}\label{fig:farsegpp}
\end{subfigure}
\begin{subfigure}{0.13\textwidth}
\includegraphics[width=\textwidth] {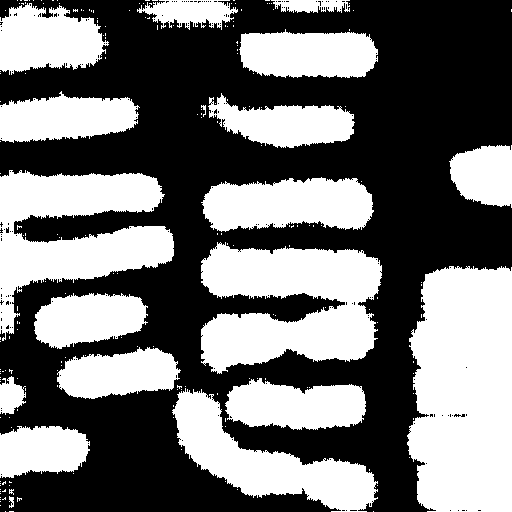}\caption*{DSMNet}\label{fig:dsmnet}
\end{subfigure}
\begin{subfigure}{0.13\textwidth}
\includegraphics[width=\textwidth] {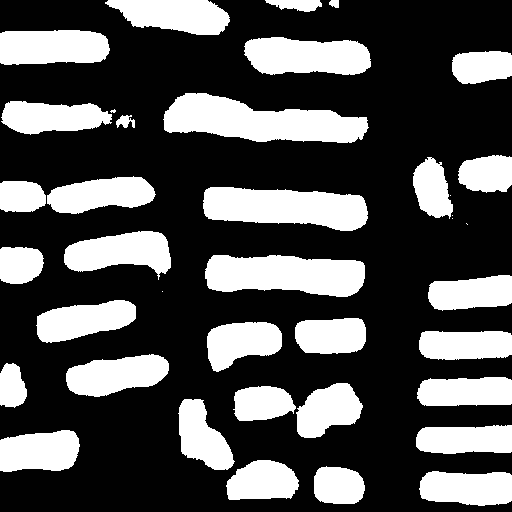}\caption*{ST-UNet}\label{fig:stunet}
\end{subfigure}
\begin{subfigure}{0.13\textwidth}
\includegraphics[width=\textwidth] {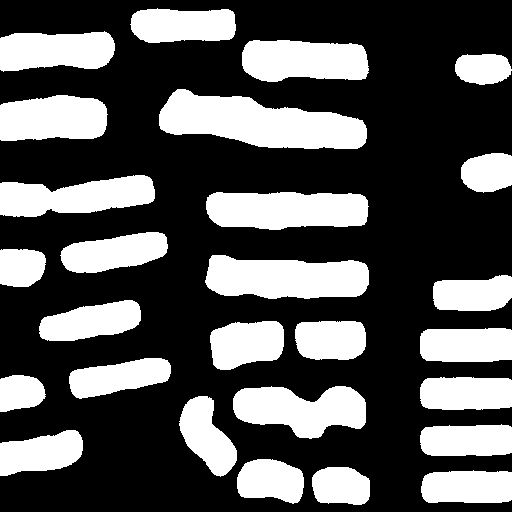}\caption*{\textbf{BuildMamba}}\label{fig:buildmamba}
\end{subfigure}

\caption{Visual comparison of segmentation results on Huawei BHE dataset.}
\label{fig:visualseghuw}
\end{figure*}

\textbf{Height estimation results.} To assess the precision of our proposed multi-task architecture in reconstructing 3D urban geometry, we evaluate the height estimation performance of BuildMamba against five state-of-the-art baselines: LUMNet \cite{lumnet}, DSMNet \cite{dsmnet}, IM2HEIGHT \cite{IM2HEIGHT}, IM2ELEVATION \cite{IM2ELEVATION}, and HTC-DC \cite{chen2023htc}. As detailed in Table~\ref{tab:height_results_comparison}, BuildMamba establishes a new performance upper bound for RGB-only methods across all three heterogeneous datasets, consistently achieving the lowest error rates and highest accuracy percentages (Note that, LUMNet utilizes building masks as auxiliary input for height estimation and is provided for benchmark purposes only). Specifically, our model achieves a mean RMSE of 1.058~m on DFC19  and 1.772~m on DFC23 datasets, representing a significant reduction in error compared to the previous best-performing models. Notably, on the DFC23 benchmark, BuildMamba outperforms the runner-up, HTC-DC, by a substantial margin of approximately 31.6\% reduction in RMSE while simultaneously attaining a $\delta_1$ accuracy of 0.881. The robustness of our architecture is further evidenced by its performance on the complex Huawei BHE dataset; while established models like IM2HEIGHT  and IM2ELEVATION  exhibit limited scalability—yielding high error rates and lower $\delta_n$ thresholds—BuildMamba maintains a competitive RMSE of 9.230~m and a high $\delta_1$ score of 0.806. HTC-DC, on the other hand, fails to deal with the inconsistent and missing height values in the dataset, yielding substantially lower $\delta_n$ percentages. Furthermore, while models utilizing privileged information like LUMNet (RGB+Mask) are inherently more powerful due to the additional structural constraints provided by ground truth segmentation masks, BuildMamba achieves highly comparable results using only RGB input, underscoring the effectiveness of the model's advanced feature representation of both local and global context. 

These quantitative results demonstrate that the selective scanning mechanism of the Mamba-based BuildMamba model effectively captures the long-range spatial dependencies and depth cues required for accurate height inference, confirming that our joint learning strategy successfully optimizes both semantic and geometric feature representations.

\subsection{Qualitative Results}
\textbf{Segmentation results.}
We evaluate the performance of our proposed model through a qualitative comparison on the DFC19 and Huawei BHE datasets. As illustrated in Figure \ref{fig:visualseg} for DFC19, BuildMamba produces more detailed and consistent segmentation masks, when compared to outputs of UANet \cite{uanet}, BuildFormer \cite{BuildFormer}, FarSeg++ \cite{FarSeg++}, DSMNet \cite{dsmnet} and ST-UNet \cite{ST-Unet}. Specifically, our model produces significantly cleaner predictions with substantially fewer false positives and false negatives compared to the other architectures. Furthermore, the building contours in our results are noticeably sharper and more defined, maintaining a clear separation between individual masks, mirroring closely the ground truth maps. This precision in edge delineation and the reduction of false positives/negatives demonstrate the effectiveness of our multi-task strategy in capturing complex building footprints.

As illustrated in Figure \ref{fig:visualseghuw} for Huawei dataset, our model exhibits superior qualitative results in delineating complex building footprints compared to existing segmentation baselines. While convolutional and transformer-based models like UANet, BuildFormer, and ST-UNet often produce irregular boundaries or eroded building masks, BuildMamba produces fewer false negatives, geometrically more consistent masks and clear separation between closely situated structures. Notably, the model effectively captures the geometric orientation and varied shapes of the footprints, avoiding the over-smoothing seen in FarSeg++ and the significant noise present in the DSMNet output. This visual precision indicates that the state-space modeling within BuildMamba is highly adept at capturing fine-grained local details while maintaining global context, allowing it to accurately segment rows of buildings even under varying lighting conditions and complex textures. By consistently producing masks that closely mirror the ground truth, BuildMamba proves that its architectural benefits extend beyond height estimation to highly accurate semantic segmentation in challenging urban environments.

\begin{figure*}[t]
\centering

\begin{subfigure}{0.15\textwidth}
\includegraphics[width=\textwidth] {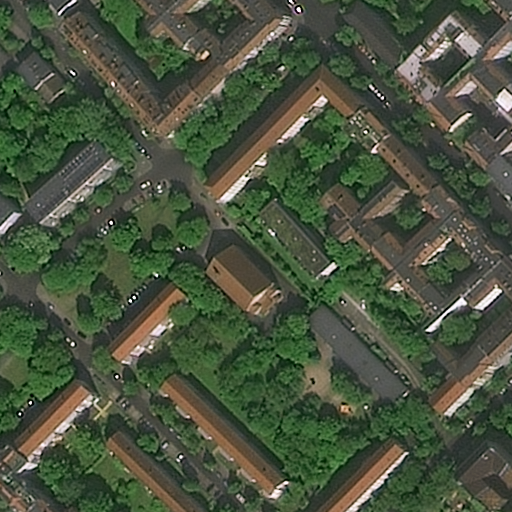}\caption*{RGB  }\label{fig:rgb_height_dfc23}
\end{subfigure}
\begin{subfigure}{0.15\textwidth}
\includegraphics[width=\textwidth] {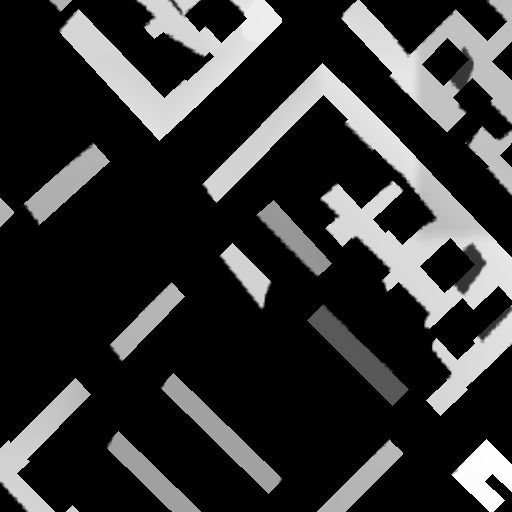}\caption*{Ground Truth}\label{fig:gt_height_dfc23}
\end{subfigure}
\begin{subfigure}{0.15\textwidth}
\includegraphics[width=\textwidth] {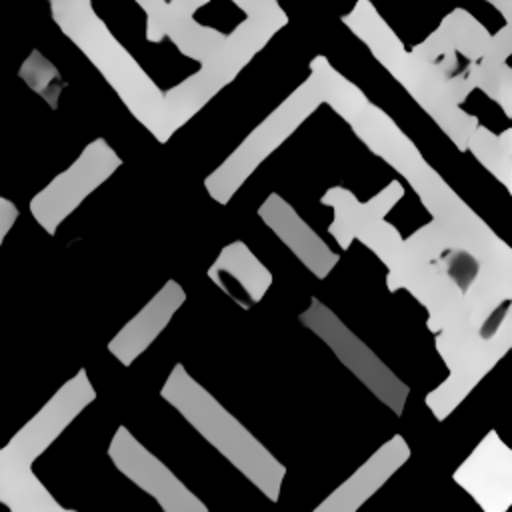}\caption*{HTC-DC}\label{fig:htcdcdfc23}
\end{subfigure}

\begin{subfigure}{0.15\textwidth}
\includegraphics[width=\textwidth] {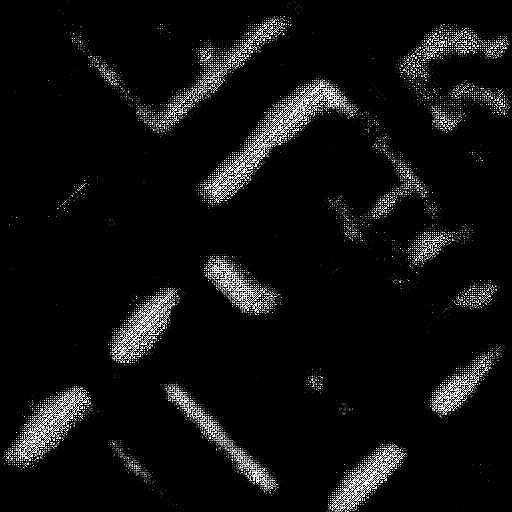}\caption*{IM2HEIGHT}\label{fig:im2heightdfc23}
\end{subfigure}
\begin{subfigure}{0.15\textwidth}
\includegraphics[width=\textwidth] {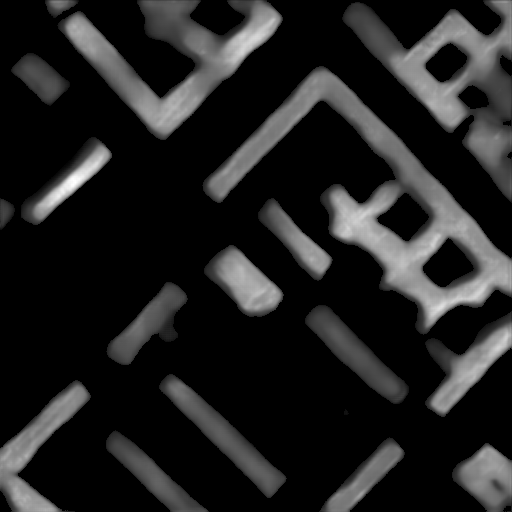}\caption*{DSMNet}\label{fig:heightdfc23dsmnet}
\end{subfigure}
\begin{subfigure}{0.15\textwidth}
\includegraphics[width=\textwidth] {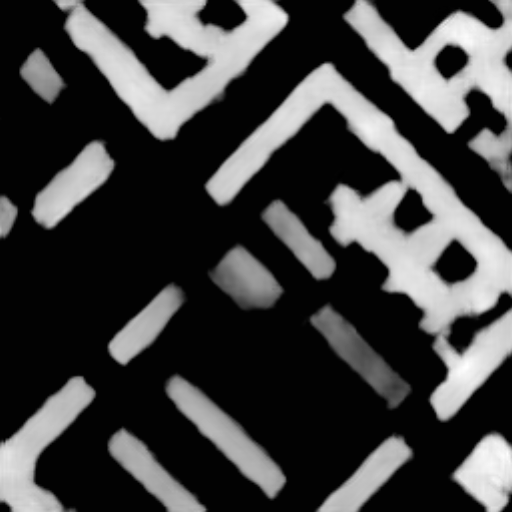}\caption*{LUMNet}\label{fig:heightdfc23lumnet}
\end{subfigure}
\begin{subfigure}{0.15\textwidth}
\includegraphics[width=\textwidth] {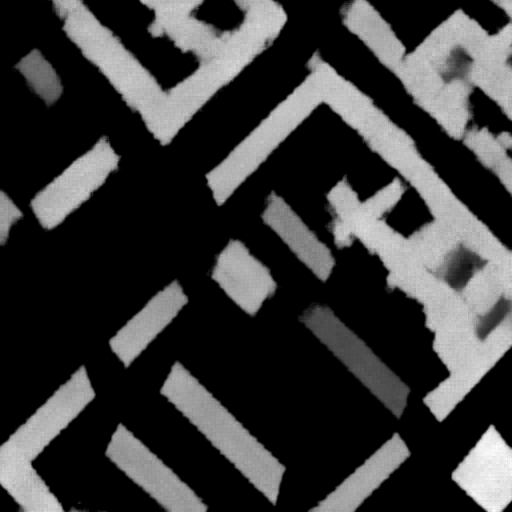}\caption*{\textbf{BuildMamba}}\label{fig:heightdfc23buildmamba}
\end{subfigure}

\caption{Visual comparison of height estimation results DFC23 dataset}
\label{fig:visualheightdfc23}
\end{figure*}

\begin{figure*}[t]
\centering

\begin{subfigure}{0.15\textwidth}
\includegraphics[width=\textwidth] {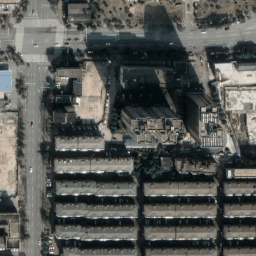}\caption*{RGB  }\label{fig:rgb_height_huawei}
\end{subfigure}
\begin{subfigure}{0.15\textwidth}
\includegraphics[width=\textwidth] {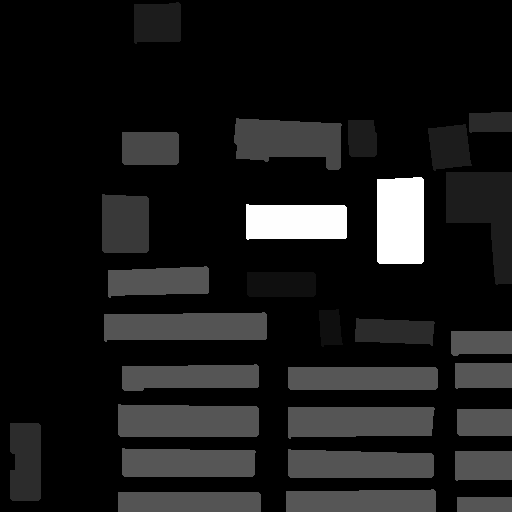}\caption*{Ground Truth}\label{fig:gt_height_huawei}
\end{subfigure}
\begin{subfigure}{0.15\textwidth}
\includegraphics[width=\textwidth] {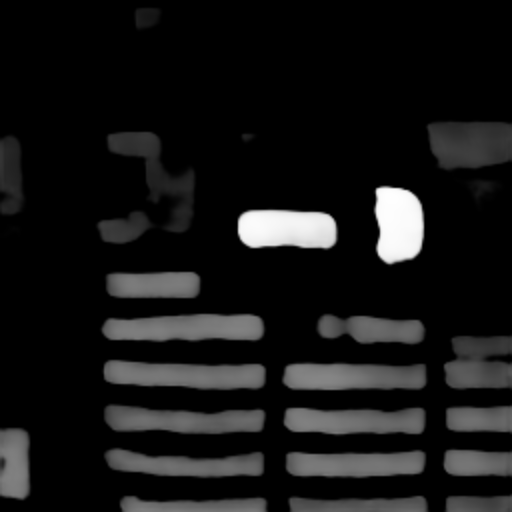}\caption*{HTC-DC}\label{fig:htcdcdfc23}
\end{subfigure}

\begin{subfigure}{0.15\textwidth}
\includegraphics[width=\textwidth] {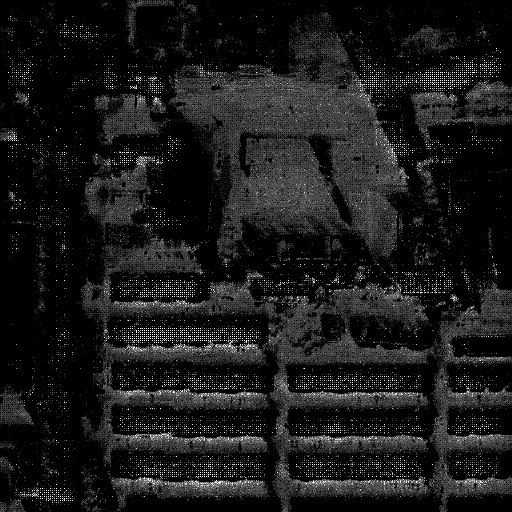}\caption*{IM2HEIGHT}\label{fig:im2heightdfc23}
\end{subfigure}
\begin{subfigure}{0.15\textwidth}
\includegraphics[width=\textwidth] {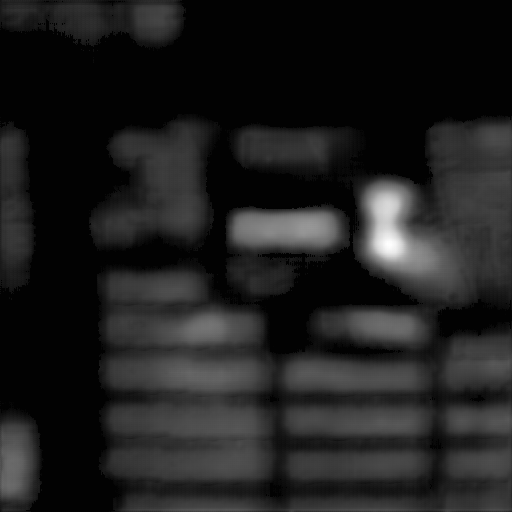}\caption*{DSMNet}\label{fig:heighthuaweidsmnet}
\end{subfigure}
\begin{subfigure}{0.15\textwidth}
\includegraphics[width=\textwidth] {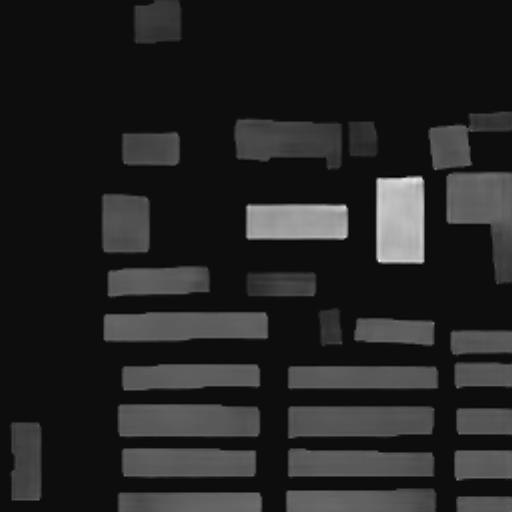}\caption*{LUMNet}\label{fig:heighthuaweilumnet}
\end{subfigure}
\begin{subfigure}{0.15\textwidth}
\includegraphics[width=\textwidth] {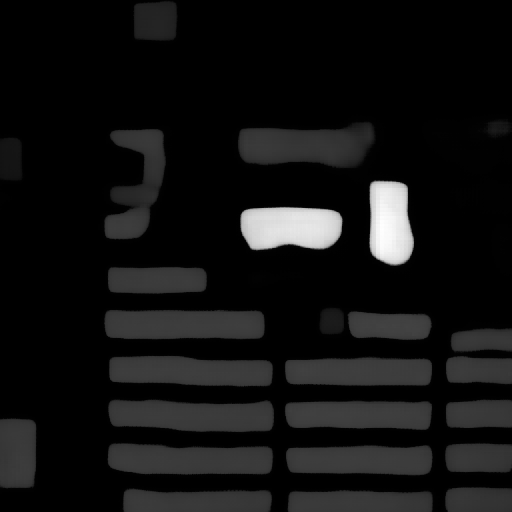}\caption*{\textbf{BuildMamba}}\label{fig:heighthuaweibuildmamba}
\end{subfigure}

\caption{Visual comparison of height estimation results on Huawei BHE dataset}
\label{fig:visualheighthuawei}
\end{figure*}

\textbf{Height estimation results.}
In this section, we provide a visual assessment of height estimation by BuildMamba to evaluate its structural fidelity and edge preservation capabilities. 

The qualitative results on the DFC23 dataset, illustrated in Figure \ref{fig:visualheightdfc23}, visually substantiate the superior performance of BuildMamba by demonstrating its ability to reconstruct sharp, geometrically precise building profiles that closely align with the ground truth. While baseline methods such as IM2HEIGHT and DSMNet suffer from severe structural blurring or artifacts, BuildMamba maintains high-fidelity edges and clear contrast between rooftops and ground levels. This visual clarity is numerically validated by the DFC23 metrics, where BuildMamba achieves a state-of-the-art RMSE of 1.772~m, an improvement of 0.82~m over the nearest RGB-only competitor, HTC-DC. The model’s success in preserving high-frequency spatial details and producing consistent height gradations across dense urban blocks highlights the effectiveness of its Mamba-based architecture in capturing long-range spatial dependencies without the need for auxiliary mask input during inference, which is a significant advancement over models like LUMNet.

Moreover, the qualitative results on the Huawei BHE dataset, depicted in Figure \ref{fig:visualheighthuawei}, further underscore the robustness of BuildMamba in complex urban environments, where it demonstrates a superior capacity for maintaining structural integrity and precise height gradations compared to existing baselines. While methods such as HTC-DC and DSMNet suffer from significant blurring and structural inaccuracies between adjacent buildings, BuildMamba produces remarkably sharp footprints that closely mirror the ground truth. This visual fidelity is supported by the quantitative metrics in the Huawei BHE evaluation, where BuildMamba achieves a $\delta_1$ accuracy of 0.806, representing a substantial leap over 0.479 achieved by HTC-DC and 0.609 by DSMNet. Even when compared to LUMNet, which utilizes additional mask information, BuildMamba delivers more accurate height estimates as long as the corresponding building footprint is accurately detected. The model's ability to capture the relative height differences of high-rise structures without the hallucinated artifacts seen in other RGB-based models validates the efficiency of its Mamba-based architecture in modeling the intricate spatial dependencies inherent in satellite imagery.

\begin{table*}[htbp]
\caption{Ablation Study on DFC19 Dataset. The best result is highlighted in bold.}
\centering
\scalebox{1}{
\begin{tabular}{cccc|ccc}
\toprule
VMamba & MAM & S-MambaFPN & MHR & IoU  & RMSE (m) & Params (M)   \\ 
\midrule
             &  &   &   &         0.785 & - & 49.2 \\ 
    \checkmark   &    &   &    &    0.852 &  1.20 & 58.2 \\ 
      \checkmark & \checkmark     &     &    & 0.871 & 1.17 & 65.2 \\
      \checkmark & \checkmark     &  \checkmark   &    &  0.896 & 1.10 & 71.2 \\
      \rowcolor{black!10}

      \checkmark & \checkmark     &  \checkmark   &   \checkmark & \textbf{0.897} & \textbf{1.06} & 74.3  \\
\bottomrule
\end{tabular}
}
\label{table:abb}
\vskip-8.0pt
\end{table*}

\subsection{Ablation Study}

To evaluate the contribution of each proposed component, we conducted a comprehensive ablation study on the DFC19 dataset, as shown in Table~\ref{table:abb}. The baseline model is BuildFormer \cite{BuildFormer}, which serves as the foundation for all variants.

Since BuildFormer \cite{BuildFormer} is natively a single-task architecture designed exclusively for building segmentation, it lacks the necessary regression heads for height estimation. Consequently,  BuildFormer is evaluated solely on the segmentation task, and no height estimation metrics are reported for this baseline.

VMamba denotes the integration of the Visual Mamba  instead of a Tranformer backbone, which introduces long-range dependency modeling through selective state-space representations. A simple height regression head is also added to the architecture at this stage. Compared to the baseline, VMamba significantly enhances segmentation accuracy, improving IoU from 0.785 to 0.852, confirming its strong capability for contextual feature modeling.

Adding the Mamba Attention Module (MAM) further refines feature interactions between global and local representations, leading to an IoU of 0.871 and an RMSE of 1.17 m. This demonstrates that attention-guided fusion effectively stabilizes feature propagation across scales.

Next, the Spatial-aware MambaFPN (S-MambaFPN) enhances hierarchical feature aggregation by selectively propagating spatially discriminative representations. With this addition, the IoU increases to 0.896 and the RMSE decreases to 1.10 m, demonstrating more accurate structural and boundary modeling.

Finally, incorporating the mask-aware height refinement (MHR) module—responsible for height estimation refinement guided by segmentation cues—yields the best performance with an IoU of 0.897 and an RMSE of 1.06 m. Since MHR focuses on height refinement, it does not alter the segmentation scores but notably improves height accuracy.

Overall, the incremental improvements across variants confirm that each proposed component synergistically contributes to more reliable multi-task height estimation and segmentation performance.

\begin{figure*}[ht]
  \centering
  \includegraphics[width=0.6\linewidth]{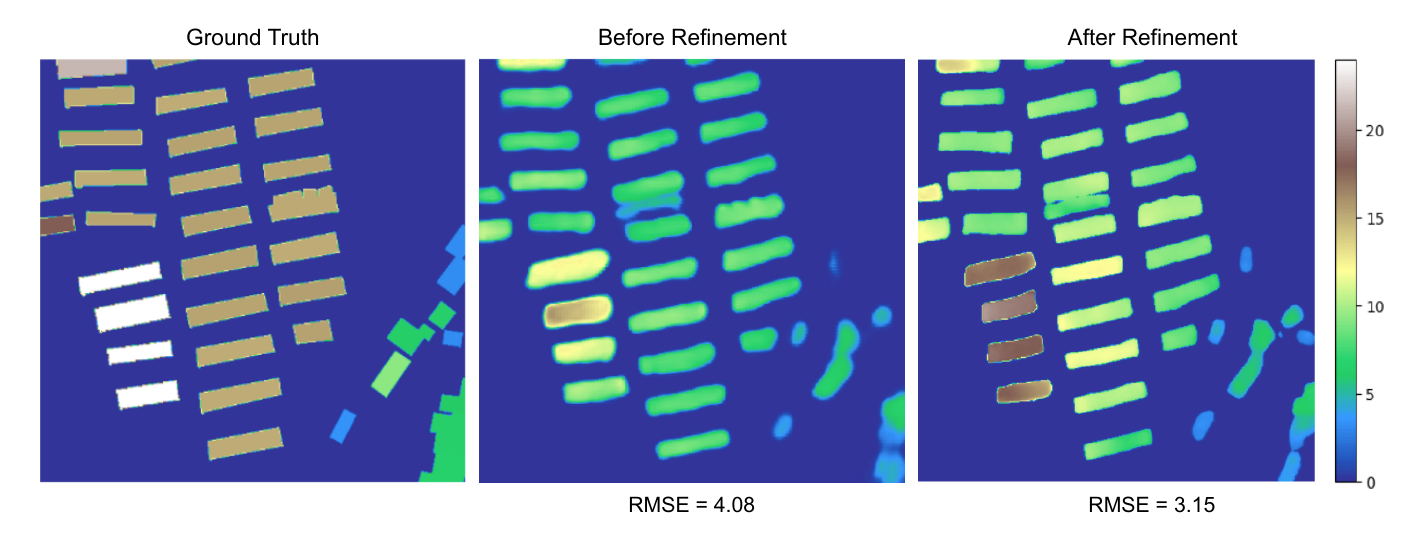}
  \caption{Qualitative analysis of the MHR module on the DFC19 dataset.}
  \label{fig:qual_refinement}
\end{figure*}

\subsubsection{Analysis of Backbone Module}

We further conduct a backbone analysis to assess the representational advantages of our Mamba-based architecture against widely used alternatives. As summarized in Table~\ref{tab:backbone}, our model is compared with two strong baselines—ResNet-101~\cite{resnet} and Swin-S~\cite{swin}—under identical training settings on both DFC19 and DFC23 datasets. The proposed Mamba backbone achieves superior performance across all metrics, lifting IoU from 0.862 to 0.897 on DFC19 and from 0.891 to 0.927 on DFC23, while consistently reducing RMSE from 1.11~m to 1.05~m and from 1.91~m to 1.80~m, respectively. These improvements highlight Mamba’s strong capacity for long-range dependency modeling and efficient spatial aggregation, enabling more contextually coherent predictions. Importantly, this performance gain comes with marginal increase in parameters (74.3M vs. 74.4M for Swin-S and 60.2M for ResNet-101), demonstrating that the proposed Mamba backbone offers a more favorable trade-off between accuracy and complexity compared to conventional CNN- and Transformer-based designs.

\begin{table}[htbp]
\centering
\renewcommand{\arraystretch}{1.1}
\caption{Analysis of backbone on DFC19 and DFC23 datasets using IoU and RMSE (m) metrics.}
\label{tab:backbone}
\resizebox{\linewidth}{!}{
\begin{tabular}{lccccc}
\toprule
            & \multicolumn{2}{c}{\textbf{DFC19}} & \multicolumn{2}{c}{\textbf{DFC23}} & \textbf{Params (M)} \\
            \cmidrule(lr){2-3}\cmidrule(lr){4-5}

            & \textbf{IoU} & \textbf{RMSE }              & \textbf{IoU} & \textbf{RMSE}                &        \\
\midrule
ResNet-101       &0.862  &1.11  &0.891  &1.91  &\textbf{60.2}        \\
Swin-S           &0.881  &1.09  &0.908  &1.85  &74.4        \\
\rowcolor{black!10}
Mamba (ours)     &\textbf{0.897}  &\textbf{1.05}  &\textbf{0.927}  &\textbf{1.80}  &74.3        \\
\bottomrule
\end{tabular}}
\end{table}

\subsubsection{Analysis of S-MambaFPN Module}
We perform a controlled ablation to quantify the contribution of S-MambaFPN to long-range context modeling, evaluating on DFC19 and DFC23 using complementary metrics—IoU and RMSE —both with and without the module; the full results are summarized in Table~\ref{tab:mambafpn}. On DFC19, S-MambaFPN lifts IoU from 0.872 to 0.897 and reduces RMSE from 1.16~m to 1.05~m, while on DFC23 it increases IoU from 0.909 to 0.927 and lowers RMSE from 1.98~m to 1.80~m. Importantly, this improvement comes with only a modest parameter increase from 68.3M to 74.3M, indicating that the added modeling capacity incurs minimal complexity overhead. Notably, the enhancement ratio is slightly larger on DFC19 across both metrics, consistent with our hypothesis that S-MambaFPN's long-range dependency capture more effectively exploits broader spatial extent and scale variability in DFC19, yielding proportionally greater gains where contextual aggregation is most demanded.

\begin{table}[t]
\centering
\renewcommand{\arraystretch}{1.1}
\caption{Analysis of S-MambaFPN on DFC19 and DFC23 datasets using IoU and RMSE (m) metrics.}
\label{tab:mambafpn}
\resizebox{\linewidth}{!}{
\begin{tabular}{lccccc}
\toprule
 & \multicolumn{2}{c}{\textbf{DFC19}} & \multicolumn{2}{c}{\textbf{DFC23}} & \textbf{Params (M)} \\
\cmidrule(lr){2-3}\cmidrule(lr){4-5}
& \textbf{IoU} & \textbf{RMSE} & \textbf{IoU} & \textbf{RMSE} \\
\midrule
Without S-MambaFPN & 0.872 & 1.16 & 0.909 & 1.98 & \textbf{68.3} \\
\rowcolor{black!10}
With S-MambaFPN          & \textbf{0.897} & \textbf{1.05} & \textbf{0.927} & \textbf{1.80} & 74.3 \\
\bottomrule
\end{tabular}}
\end{table}

\subsubsection{Analysis of MHR Module}

As seen in Table~\ref{tab:height_refiner},
the mask-aware height refinement consistently reduces error on both benchmarks. On DFC19, RMSE drops from 1.10~m to 1.05~m, and on DFC23 from 1.84~m to 1.77~m. Qualitatively, the largest gains appear in low-texture expanses (e.g., roads, roofs, water) where the baseline exhibits speckle and staircase artifacts, and around semantic boundaries where bleeding is common (see Figure~\ref{fig:qual_refinement}). These improvements align with the module’s design: a shallow residual predictor focuses on height-specific artifacts, while confidence-gated updates concentrate corrections on reliable supports indicated by the segmentation, attenuating them elsewhere to preserve valid elevations. Because the refiner operates on a two-channel input with optional dilations, it promotes within-region smoothness without eroding edges and adds negligible computational overhead. Training remains stable, and the refined maps are visually cleaner with sharper discontinuities, confirming that a confidence-aware, structure-preserving residual is an effective inductive bias for height correction across datasets.

\begin{table}[t]
\centering
\caption{Effect of the mask-aware height refinement on DFC19 and DFC23 datasets.}
\label{tab:height_refiner}
\renewcommand{\arraystretch}{1.15}
\setlength{\tabcolsep}{10pt}
\begin{tabular}{lcc}
\toprule
\textbf{Method} & \textbf{DFC19 RMSE (m)} & \textbf{DFC23 RMSE (m)} \\
\midrule
w/o Refinement          & 1.10 & 1.84 \\
\rowcolor{black!10}

w Refinement  & \textbf{1.06} & \textbf{1.77} \\
\bottomrule
\end{tabular}
\end{table}

\section{Conclusion}
This paper introduces BuildMamba, a unified multi-task framework motivated by the inherent challenges of monocular 3D urban mapping: boundary ambiguity in dense layouts, and the heavy computational cost of traditional global  modeling. To address these, we propose an architecture centered on the efficiency of visual state-space models. The core of the framework consists of three specialized modules: the Mamba Attention Module (MAM), which refines global-local feature interactions; the Spatial-Aware Mamba-FPN (S-MambaFPN), which utilizes gated state-space scans to aggregate multi-scale context without the quadratic overhead of Transformers; and the Mask-Aware Height Refinement (MHR) module, which treats the segmentation output as a structural prior to clean height speckle and sharpen vertical discontinuities. We evaluate BuildMamba on the DFC19, DFC23, and the heterogeneous Huawei BHE datasets, the latter representing a complex real-world benchmark with significant domain shifts. Our experiments demonstrate that BuildMamba consistently outperforms state-of-the-art baselines such as UANet and HTC-DC, establishing a new performance upper bound in both IoU and RMSE, while demonstrating superior robustness in predicting high-rise structures and maintaining crisp building footprints from single-view RGB imagery.

\section*{Acknowledgment}
This work is supported in part by TÜBİTAK Grant No: 5230108. We thank  Huawei Turkiye R\&D Center for providing the Huawei BHE dataset.

\ifCLASSOPTIONcaptionsoff
  \newpage
\fi

\bibliographystyle{IEEEtranN}
\bibliography{bibliography}

\begin{thebibliography}{42}
\providecommand{\natexlab}[1]{#1}
\providecommand{\url}[1]{#1}
\csname url@samestyle\endcsname
\providecommand{\newblock}{\relax}
\providecommand{\bibinfo}[2]{#2}
\providecommand{\BIBentrySTDinterwordspacing}{\spaceskip=0pt\relax}
\providecommand{\BIBentryALTinterwordstretchfactor}{4}
\providecommand{\BIBentryALTinterwordspacing}{\spaceskip=\fontdimen2\font plus
\BIBentryALTinterwordstretchfactor\fontdimen3\font minus \fontdimen4\font\relax}
\providecommand{\BIBforeignlanguage}[2]{{%
\expandafter\ifx\csname l@#1\endcsname\relax
\typeout{** WARNING: IEEEtranN.bst: No hyphenation pattern has been}%
\typeout{** loaded for the language `#1'. Using the pattern for}%
\typeout{** the default language instead.}%
\else
\language=\csname l@#1\endcsname
\fi
#2}}
\providecommand{\BIBdecl}{\relax}
\BIBdecl

\bibitem[Li et~al.(2016)Li, Stein, and Bijker]{li2016urban}
M.~Li, A.~Stein, and W.~Bijker, ``Urban land use extraction from very high resolution remote sensing images by {Bayesian} network,'' in \emph{IEEE International Geoscience and Remote Sensing Symposium (IGARSS)}, 2016, pp. 3334--3337.

\bibitem[Biljecki et~al.(2015)Biljecki, Stoter, Ledoux, Zlatanova, and Çöltekin]{ijgi4042842}
\BIBentryALTinterwordspacing
F.~Biljecki, J.~Stoter, H.~Ledoux, S.~Zlatanova, and A.~Çöltekin, ``Applications of 3d city models: State of the art review,'' \emph{ISPRS International Journal of Geo-Information}, vol.~4, no.~4, pp. 2842--2889, 2015. [Online]. Available: \url{https://www.mdpi.com/2220-9964/4/4/2842}
\BIBentrySTDinterwordspacing

\bibitem[Sirko et~al.(2021)Sirko, Kashubin, Ritter, Annkah, Bouchareb, Dauphin, Keysers, Neumann, Cisse, and Quinn]{sirko2021openbuildings}
W.~Sirko, S.~Kashubin, M.~Ritter, A.~Annkah, Y.~Bouchareb, Y.~Dauphin, D.~Keysers, M.~Neumann, M.~Cisse, and J.~Quinn, ``Continental-scale building detection from high resolution satellite imagery,'' 07 2021.

\bibitem[Schröter et~al.(2018)Schröter, Lüdtke, Redweik, Meier, Bochow, Ross, Nagel, and Kreibich]{schroeter2018flood}
\BIBentryALTinterwordspacing
K.~Schröter, S.~Lüdtke, R.~Redweik, J.~Meier, M.~Bochow, L.~Ross, C.~Nagel, and H.~Kreibich, ``Flood loss estimation using 3d city models and remote sensing data,'' \emph{Environmental Modelling \& Software}, vol. 105, pp. 118--131, 2018. [Online]. Available: \url{https://www.sciencedirect.com/science/article/pii/S1364815217308654}
\BIBentrySTDinterwordspacing

\bibitem[Labetski et~al.(2023)Labetski, Vitalis, Biljecki, Ohori, and Stoter]{urbanmorph2022}
\BIBentryALTinterwordspacing
A.~Labetski, S.~Vitalis, F.~Biljecki, K.~A. Ohori, and J.~Stoter, ``3d building metrics for urban morphology,'' \emph{International Journal of Geographical Information Science}, vol.~37, no.~1, pp. 36--67, 2023. [Online]. Available: \url{https://doi.org/10.1080/13658816.2022.2103818}
\BIBentrySTDinterwordspacing

\bibitem[Al-Hourani et~al.(2014)Al-Hourani, Kandeepan, and Jamalipour]{alhourani2014uav}
A.~Al-Hourani, S.~Kandeepan, and A.~Jamalipour, ``Modeling air-to-ground path loss for low altitude platforms in urban environments,'' in \emph{2014 IEEE Global Communications Conference}, 2014, pp. 2898--2904.

\bibitem[Oostwegel et~al.(2025)Oostwegel, Schorlemmer, and Gu{\'e}guen]{oostwegel2025openbuildingmap}
L.~J.~N. Oostwegel, D.~Schorlemmer, and P.~Gu{\'e}guen, ``From footprints to functions: A comprehensive global and semantic building footprint dataset,'' \emph{Scientific Data}, vol.~12, no.~1, p. 1699, Oct. 2025.

\bibitem[Milojevic-Dupont et~al.(2023)Milojevic-Dupont, Wagner, Nachtigall, Hu, Br{\"u}ser, Zumwald, Biljecki, Heeren, Kaack, Pichler, and Creutzig]{eubucco2023}
N.~Milojevic-Dupont, F.~Wagner, F.~Nachtigall, J.~Hu, G.~B. Br{\"u}ser, M.~Zumwald, F.~Biljecki, N.~Heeren, L.~H. Kaack, P.-P. Pichler, and F.~Creutzig, ``{EUBUCCO} v0.1: European building stock characteristics in a common and open database for 200+ million individual buildings,'' \emph{Scientific Data}, vol.~10, no.~1, p. 147, Mar. 2023.

\bibitem[Mou and Zhu(2018)]{IM2HEIGHT}
L.~Mou and X.~X. Zhu, ``Im2height: Height estimation from single monocular imagery via fully residual convolutional-deconvolutional network,'' \emph{arXiv preprint arXiv:1802.10249}, 2018.

\bibitem[Liu et~al.(2020)Liu, Krylov, Kane, Kavanagh, and Dahyot]{IM2ELEVATION}
C.-J. Liu, V.~A. Krylov, P.~Kane, G.~Kavanagh, and R.~Dahyot, ``Im2elevation: Building height estimation from single-view aerial imagery,'' \emph{remote sensing}, vol.~12, no.~17, p. 2719, 2020.

\bibitem[Wang et~al.(2022)Wang, Fang, Meng, and Li]{BuildFormer}
L.~Wang, S.~Fang, X.~Meng, and R.~Li, ``Building extraction with vision transformer,'' \emph{IEEE Transactions on Geoscience and Remote Sensing}, vol.~60, pp. 1--11, 2022.

\bibitem[Fu et~al.(2018)Fu, Gong, Wang, Batmanghelich, and Tao]{dorn}
H.~Fu, M.~Gong, C.~Wang, K.~Batmanghelich, and D.~Tao, ``Deep ordinal regression network for monocular depth estimation,'' in \emph{Proceedings of the IEEE conference on computer vision and pattern recognition}, 2018, pp. 2002--2011.

\bibitem[Bhat et~al.(2021)Bhat, Alhashim, and Wonka]{Adabins}
S.~F. Bhat, I.~Alhashim, and P.~Wonka, ``Adabins: Depth estimation using adaptive bins,'' in \emph{Proceedings of the IEEE/CVF conference on computer vision and pattern recognition}, 2021, pp. 4009--4018.

\bibitem[Li et~al.(2024{\natexlab{a}})Li, Wang, Liu, and Jiang]{binsformer}
Z.~Li, X.~Wang, X.~Liu, and J.~Jiang, ``Binsformer: Revisiting adaptive bins for monocular depth estimation,'' \emph{IEEE Transactions on Image Processing}, 2024.

\bibitem[Chen et~al.(2023{\natexlab{a}})Chen, Shi, Xiong, and Zhu]{chen2023htc}
S.~Chen, Y.~Shi, Z.~Xiong, and X.~X. Zhu, ``Htc-dc net: Monocular height estimation from single remote sensing images,'' \emph{IEEE Transactions on Geoscience and Remote Sensing}, vol.~61, pp. 1--18, 2023.

\bibitem[Chen et~al.(2023{\natexlab{b}})Chen, Yan, and Huang]{mftsc}
Y.~Chen, Q.~Yan, and W.~Huang, ``Mftsc: A semantically constrained method for urban building height estimation using multiple source images,'' \emph{Remote Sensing}, vol.~15, no.~23, p. 5552, 2023.

\bibitem[Lu et~al.(2023{\natexlab{a}})Lu, Jiao, Liu, Li, Liu, Liu, and Yang]{tridentmtl}
X.~Lu, L.~Jiao, Q.~Liu, L.~Li, F.~Liu, X.~Liu, and Y.~Yang, ``Trident cooperation network for building extraction and height estimation,'' in \emph{IEEE International Geoscience and Remote Sensing Symposium (IGARSS)}.\hskip 1em plus 0.5em minus 0.4em\relax IEEE, 2023, pp. 762--765.

\bibitem[Jamal and Aribisala(2023)]{sardatafusion}
S.~A. Jamal and A.~Aribisala, ``Data fusion for multi-task learning of building extraction and height estimation,'' \emph{arXiv preprint arXiv:2308.02960}, 2023.

\bibitem[Du et~al.(2024)Du, Xing, Wang, Xiao, Li, and Liu]{lumnet}
S.~Du, J.~Xing, S.~Wang, X.~Xiao, J.~Li, and H.~Liu, ``Lumnet: Land use knowledge guided multiscale network for height estimation from single remote sensing images,'' \emph{IEEE Geoscience and Remote Sensing Letters}, vol.~21, pp. 1--5, 2024.

\bibitem[Zheng et~al.(2019)Zheng, Zhong, and Wang]{popnet}
Z.~Zheng, Y.~Zhong, and J.~Wang, ``Pop-net: Encoder-dual decoder for semantic segmentation and single-view height estimation,'' in \emph{IEEE International Geoscience and Remote Sensing Symposium (IGARSS)}.\hskip 1em plus 0.5em minus 0.4em\relax IEEE, 2019, pp. 4963--4966.

\bibitem[Elhousni et~al.(2021)Elhousni, Zhang, and Huang]{dsmnet}
M.~Elhousni, Z.~Zhang, and X.~Huang, ``Height prediction and refinement from aerial images with semantic and geometric guidance,'' \emph{IEEE Access}, vol.~9, pp. 145\,638--145\,647, 2021.

\bibitem[He et~al.(2022)He, Zhou, Zhao, Zhang, Yao, and Xue]{ST-Unet}
X.~He, Y.~Zhou, J.~Zhao, D.~Zhang, R.~Yao, and Y.~Xue, ``Swin transformer embedding unet for remote sensing image semantic segmentation,'' \emph{IEEE transactions on geoscience and remote sensing}, vol.~60, pp. 1--15, 2022.

\bibitem[Zheng et~al.(2023)Zheng, Zhong, Wang, Ma, and Zhang]{FarSeg++}
Z.~Zheng, Y.~Zhong, J.~Wang, A.~Ma, and L.~Zhang, ``Farseg++: Foreground-aware relation network for geospatial object segmentation in high spatial resolution remote sensing imagery,'' \emph{IEEE Transactions on Pattern Analysis and Machine Intelligence}, vol.~45, no.~11, pp. 13\,715--13\,729, 2023.

\bibitem[Li et~al.(2021)Li, He, Li, Li, Cheng, Shi, Weng, Tong, and Lin]{PointFlow}
X.~Li, H.~He, X.~Li, D.~Li, G.~Cheng, J.~Shi, L.~Weng, Y.~Tong, and Z.~Lin, ``Pointflow: Flowing semantics through points for aerial image segmentation,'' in \emph{Proceedings of the IEEE/CVF Conference on Computer Vision and Pattern Recognition}, 2021, pp. 4217--4226.

\bibitem[Li et~al.(2024{\natexlab{b}})Li, He, Cao, Zhang, and Zhang]{uanet}
J.~Li, W.~He, W.~Cao, L.~Zhang, and H.~Zhang, ``Uanet: An uncertainty-aware network for building extraction from remote sensing images,'' \emph{IEEE Transactions on Geoscience and Remote Sensing}, vol.~62, pp. 1--13, 2024.

\bibitem[Mao et~al.(2023)Mao, Sun, Huang, and Chen]{light}
Y.~Mao, X.~Sun, X.~Huang, and K.~Chen, ``Light: Joint individual building extraction and height estimation from satellite images through a unified multitask learning network,'' in \emph{IEEE International Geoscience and Remote Sensing Symposium (IGARSS)}.\hskip 1em plus 0.5em minus 0.4em\relax IEEE, 2023, pp. 5320--5323.

\bibitem[Liu et~al.(2024)Liu, Tian, Zhao, Yu, Xie, Wang, Ye, Jiao, and Liu]{vmamba}
Y.~Liu, Y.~Tian, Y.~Zhao, H.~Yu, L.~Xie, Y.~Wang, Q.~Ye, J.~Jiao, and Y.~Liu, ``Vmamba: Visual state space model,'' \emph{Advances in neural information processing systems}, vol.~37, pp. 103\,031--103\,063, 2024.

\bibitem[Dosovitskiy et~al.(2020)Dosovitskiy, Beyer, Kolesnikov, Weissenborn, Zhai, Unterthiner, Dehghani, Minderer, Heigold, Gelly, et~al.]{vit}
A.~Dosovitskiy, L.~Beyer, A.~Kolesnikov, D.~Weissenborn, X.~Zhai, T.~Unterthiner, M.~Dehghani, M.~Minderer, G.~Heigold, S.~Gelly \emph{et~al.}, ``An image is worth 16x16 words: Transformers for image recognition at scale,'' \emph{arXiv preprint arXiv:2010.11929}, 2020.

\bibitem[Liu et~al.(2021)Liu, Lin, Cao, Hu, Wei, Zhang, Lin, and Guo]{swin}
Z.~Liu, Y.~Lin, Y.~Cao, H.~Hu, Y.~Wei, Z.~Zhang, S.~Lin, and B.~Guo, ``Swin transformer: Hierarchical vision transformer using shifted windows,'' in \emph{Proceedings of the IEEE/CVF international conference on computer vision}, 2021, pp. 10\,012--10\,022.

\bibitem[Gao et~al.(2023)Gao, Sun, Lu, Zhang, Song, Zhang, and Zhai]{jointcontrastive}
Z.~Gao, W.~Sun, Y.~Lu, Y.~Zhang, W.~Song, Y.~Zhang, and R.~Zhai, ``Joint learning of semantic segmentation and height estimation for remote sensing image leveraging contrastive learning,'' \emph{IEEE Transactions on Geoscience and Remote Sensing}, vol.~61, pp. 1--15, 2023.

\bibitem[Ulu et~al.(2025)Ulu, Yagmur, Ates, Gunturk, and Hanoglu]{buildmamba_conference}
S.~U. Ulu, I.~C. Yagmur, H.~F. Ates, B.~K. Gunturk, and O.~Hanoglu, ``Visual state-space based multi-task learning for building segmentation and height estimation,'' in \emph{IEEE International Geoscience and Remote Sensing Symposium (IGARSS)}, 2025, pp. 8219--8223.

\bibitem[Huang et~al.(2022)Huang, Ren, Liu, Wang, Yu, Schmitt, H{\"a}nsch, Sun, Huang, and Mayer]{dfc23}
X.~Huang, L.~Ren, C.~Liu, Y.~Wang, H.~Yu, M.~Schmitt, R.~H{\"a}nsch, X.~Sun, H.~Huang, and H.~Mayer, ``Urban building classification (ubc)-a dataset for individual building detection and classification from satellite imagery,'' in \emph{Proceedings of the IEEE/CVF Conference on Computer Vision and Pattern Recognition}, 2022, pp. 1413--1421.

\bibitem[Bosch et~al.(2019)Bosch, Foster, Christie, Wang, Hager, and Brown]{dfc19}
M.~Bosch, K.~Foster, G.~Christie, S.~Wang, G.~D. Hager, and M.~Brown, ``Semantic stereo for incidental satellite images,'' in \emph{2019 IEEE Winter Conference on Applications of Computer Vision (WACV)}.\hskip 1em plus 0.5em minus 0.4em\relax IEEE, 2019, pp. 1524--1532.

\bibitem[Xu et~al.(2024)Xu, Feng, Wan, Xie, Feng, Zhu, and Liu]{shadowbuilding}
W.~Xu, Z.~Feng, Q.~Wan, Y.~Xie, D.~Feng, J.~Zhu, and Y.~Liu, ``Building height extraction from high-resolution single-view remote sensing images using shadow and side information,'' \emph{IEEE Journal of Selected Topics in Applied Earth Observations and Remote Sensing}, 2024.

\bibitem[Byrnside~III(2022)]{shadow}
L.~L. Byrnside~III, ``Shadow-based automatic building height estimation from high spatial resolution satellite imagery,'' \emph{Graduate Theses/Dissertations. 3768}, 2022.

\bibitem[He et~al.(2016)He, Zhang, Ren, and Sun]{resnet}
K.~He, X.~Zhang, S.~Ren, and J.~Sun, ``Deep residual learning for image recognition,'' in \emph{Proceedings of the IEEE conference on computer vision and pattern recognition}, 2016, pp. 770--778.

\bibitem[Lin et~al.(2017)Lin, Doll{\'a}r, Girshick, He, Hariharan, and Belongie]{fpn}
T.-Y. Lin, P.~Doll{\'a}r, R.~Girshick, K.~He, B.~Hariharan, and S.~Belongie, ``Feature pyramid networks for object detection,'' in \emph{Proceedings of the IEEE conference on computer vision and pattern recognition}, 2017, pp. 2117--2125.

\bibitem[Zheng et~al.(2020)Zheng, Zhong, Wang, and Ma]{FarSeg}
Z.~Zheng, Y.~Zhong, J.~Wang, and A.~Ma, ``Foreground-aware relation network for geospatial object segmentation in high spatial resolution remote sensing imagery,'' in \emph{Proceedings of the IEEE/CVF conference on computer vision and pattern recognition}, 2020, pp. 4096--4105.

\bibitem[Mahmud et~al.(2020)Mahmud, Price, Bapat, and Frahm]{boundaryaware}
J.~Mahmud, T.~Price, A.~Bapat, and J.-M. Frahm, ``Boundary-aware 3d building reconstruction from a single overhead image,'' in \emph{Proceedings of the IEEE/CVF Conference on Computer Vision and Pattern Recognition}, 2020, pp. 441--451.

\bibitem[Lu et~al.(2023{\natexlab{b}})Lu, Cao, Zhang, Liu, Peng, Liu, Yuan, Zhang, Huang, and Wang]{hgdnet}
C.~Lu, N.~Cao, P.~Zhang, T.~Liu, B.~Peng, G.~Liu, M.~Yuan, S.~Zhang, S.~Huang, and T.~Wang, ``Hgdnet: A height-hierarchy guided dual-decoder network for single view building extraction and height estimation,'' in \emph{IEEE International Geoscience and Remote Sensing Symposium (IGARSS)}.\hskip 1em plus 0.5em minus 0.4em\relax IEEE, 2023, pp. 758--761.

\bibitem[Kirillov et~al.(2023)Kirillov, Mintun, Ravi, Mao, Rolland, Gustafson, Xiao, Whitehead, Berg, Lo, et~al.]{sam}
A.~Kirillov, E.~Mintun, N.~Ravi, H.~Mao, C.~Rolland, L.~Gustafson, T.~Xiao, S.~Whitehead, A.~C. Berg, W.-Y. Lo \emph{et~al.}, ``Segment anything,'' in \emph{Proceedings of the IEEE/CVF international conference on computer vision}, 2023, pp. 4015--4026.

\bibitem[Doruk et~al.(2024)Doruk, Oztop, and Ates]{doruk2024transadapter}
A.~Doruk, E.~Oztop, and H.~F. Ates, ``Transadapter: Vision transformer for feature-centric unsupervised domain adaptation,'' \emph{arXiv preprint arXiv:2412.04073}, 2024.

\end{thebibliography}

\begin{IEEEbiography}[{\includegraphics[width=1in,height=1.25in,clip,keepaspectratio]{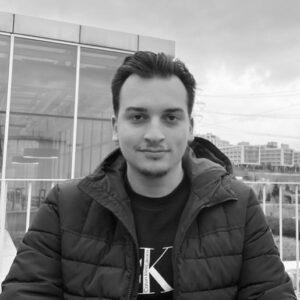}}]{SINAN U. ULU} 
received the B.S.  and M.S. degrees in  Electrical and Electronics Engineering from Ozyegin University, Istanbul, Türkiye. 
He is currently a Machine Learning Engineer at DHL Supply Chain. His research interests include remote sensing, computer vision and deep learning.
\end{IEEEbiography}

\begin{IEEEbiography}[{\includegraphics[width=1in,height=1.25in,clip,keepaspectratio]{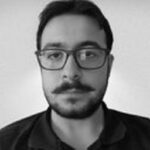}}]{A. ENES DORUK} (Student Member, IEEE)
received his B.S. degree in Electrical and Electronics Engineering from Bursa Technical University in 2022. He worked as an AI Robotic Software Engineer at DAIMIA Engineering and later as a Perception and Detection Software Engineer at ADASTEC Corporation. He is currently an M.S.student in Artificial Intelligence Engineering at Ozyegin University. His research interests include remote sensing, 3D computer vision, unsupervised domain adaptation, and multi-sensor fusion for autonomous driving.
\end{IEEEbiography}

\begin{IEEEbiography}[{\includegraphics[width=1in,height=1.25in,clip,keepaspectratio]{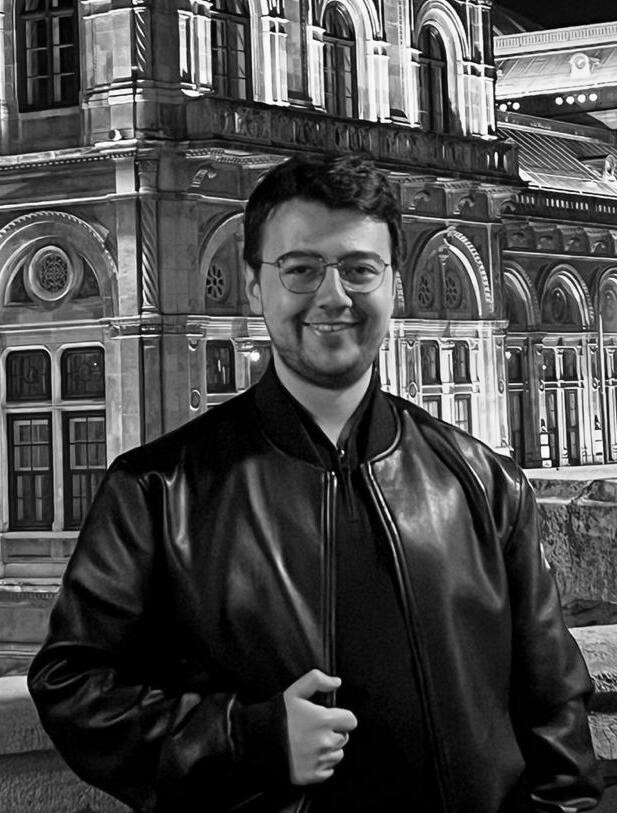}}]{I. CAN YAGMUR} (Student Member, IEEE)
received the B.S. degrees in Computer Science and Electrical and Electronics Engineering from Ozyegin University, Istanbul, Türkiye. He also received the M.S. degree in Computer Science from Ozyegin University. He is currently pursuing the Ph.D. degree in Computer Science at the University of Rochester, Rochester, NY, USA. His research interests include computational imaging, multispectral and multimodal image matching, uncertainty quantification in deep learning systems.
\end{IEEEbiography}

\begin{IEEEbiography}[{\includegraphics[width=1in,height=1.25in,clip,keepaspectratio]{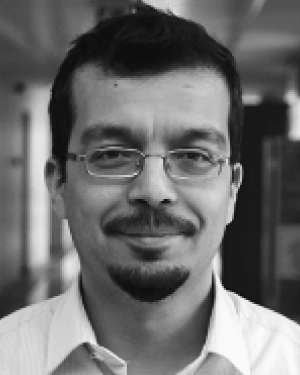}}]{BAHADIR K. GUNTURK} received the B.S. degree
in Electrical Engineering from Bilkent University,
Türkiye, in 1999, and the Ph.D. degree in Electrical
Engineering from the Georgia Institute of Technology, in 2003. From 2003 to 2014, he has been with the Department of Electrical and Computer Engineering, Louisiana State University. Since 2014,
he has been with Istanbul Medipol University,
where he is currently a Professor. He has published
more than 100 peer-reviewed journals/conference
papers in the areas of image processing and computer vision.
\end{IEEEbiography}

\begin{IEEEbiography}[{\includegraphics[width=1in,height=1.25in,clip,keepaspectratio]{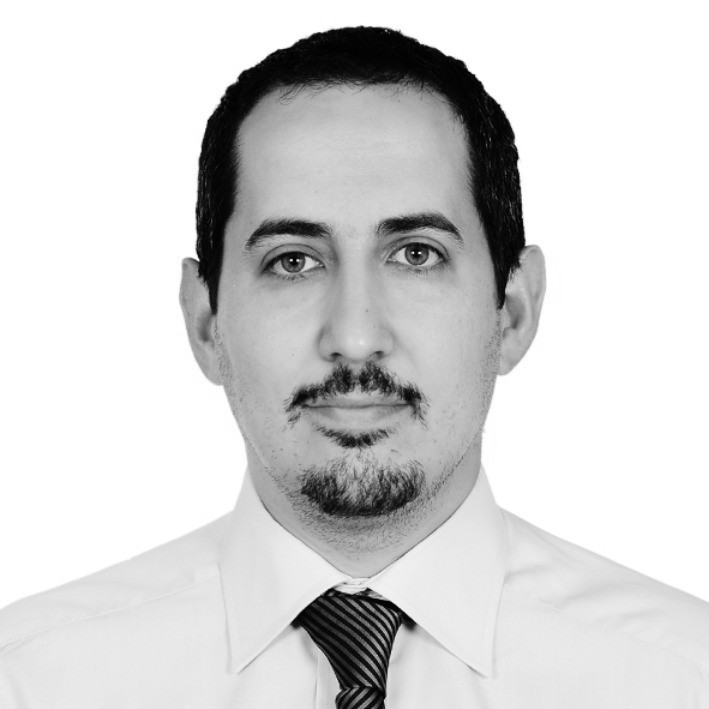}}]{OGUZ HANOGLU} received the B.S. and M.S. degrees in Electrical and Electronics engineering from Middle East Technical University and Bilkent University, Türkiye, respectively. He is currently a Computer Vision Research Engineer with the Huawei Turkey R\&D Center. His research interests include computer vision and deep learning. He was a recipient of the IEEE Electron Devices Society Masters Student Fellowship. 
\end{IEEEbiography}

\begin{IEEEbiography}[{\includegraphics[width=1in,height=1.25in,clip,keepaspectratio]{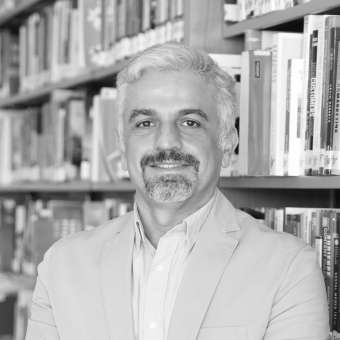}}]{HASAN F. ATES} (Senior Member, IEEE) received the Ph.D. degree from the Department of Electrical Engineering, Princeton University, in 2004.
He was a Research Associate with Sabanci University, from 2004 to 2005. He held positions of an Assistant, an Associate, and a Full Professorship
with Isik University, from 2005 to 2018. He was with Istanbul Medipol University, from 2018
to 2022. Since September 2022, he has been a Professor in the Department of Artificial Intelligence and Data Engineering, Özyeğin University. He is the author/coauthor of more than 100 peer-reviewed publications in the areas of image/video processing/coding
and computer vision.
\end{IEEEbiography}

\end{document}